\documentclass[]{interact}

\usepackage{epstopdf}
\usepackage[caption=false]{subfig}

\usepackage{caption}
\usepackage{cite}
\usepackage{lineno,hyperref}
\usepackage{multirow}
\usepackage{booktabs}
\usepackage{threeparttable}
\usepackage{bm}
\hypersetup{pdfauthor=author}

\usepackage[ruled]{algorithm2e}  
\usepackage{amssymb,amsmath}
\modulolinenumbers[5]

\usepackage[numbers,sort&compress]{natbib}
\bibpunct[, ]{[}{]}{,}{n}{,}{,}

\theoremstyle{plain}

\theoremstyle{definition}

\theoremstyle{remark}

\begin{document}

	\title{VQA and Visual Reasoning: An Overview of Recent Datasets, Methods and Challenges
	}
	
	\author{
		\name{Rufai Yusuf Zakari\textsuperscript{1}\thanks{CONTACT Rufai Yusuf Zakari. Email: rufaig6@gmail.com}, Jim Wilson Owusu\textsuperscript{2}, Hailin Wang\textsuperscript{3}, Ke Qin\textsuperscript{4}, Zaharaddeen Karami Lawal\textsuperscript{5} Yuezhou Dong\textsuperscript{6}}
		\affil{University of Electronic Science and Technology of China }\affil{Universiti Brunei Darussalam}
		{rufaig6@gmail.com\textsuperscript{1}, owilsonjim@gmail.com\textsuperscript{2}, lynn$\_$whl@msn.com\textsuperscript{3}, }qinke@uestc.edu.cn\textsuperscript{4}, 
		deenklawal13@gmail.com \textsuperscript{5}, qq1002132640@gmail.com\textsuperscript{6} }

	\maketitle
	
	\begin{abstract}
		Artificial Intelligence (AI) and its applications have sparked extraordinary interest in recent years. This achievement can be ascribed in part to advances in AI subfields including Machine Learning (ML), Computer Vision (CV), and Natural Language Processing (NLP). Deep learning, a sub-field of machine learning that employs artificial neural network concepts, has enabled the most rapid growth in these domains. The integration of vision and language has sparked a lot of attention as a result of this. The tasks have been created in such a way that they properly exemplify the concepts of deep learning. In this review paper, we provide a thorough and an extensive review of the state of the arts approaches, key models design principles and discuss existing datasets, methods, their problem formulation and evaluation measures  for VQA and Visual reasoning tasks to understand vision and language representation learning. We also present some potential future paths in this field of research, with the hope that our study may generate new ideas and novel approaches to handle existing difficulties and develop new applications.
		
	\end{abstract}
	
	\section{Introduction}
	The recent advancements of deep neural networks (DNNs) have boosted research in many fields of Artificial intelligence (AI) such as Natural language processing (NLP) and Computer vision (CV). With the exponential computational resources and the rise in dataset sizes, DNNs models such as Convolutional Neural Networks (CNNs)\cite{lecun1995convolutional}, recurrent neural networks (RNNs)\cite{hochreiter1997long}, and auto-encoders\cite{vincent2010stacked} achieved a great victory in many machine learning (ML) tasks such as object detection\cite{ren2016faster}, machine translation\cite{luong2015effective}, image caption generation\cite{donahue2015long}, speech recognition\cite{hinton2012deep}. There is, nevertheless, some curiosity in tackling problems that mix semantic and visual data from two usually separate domains. Methods for addressing the integration problem should enable comprehensive knowledge of visual or textual information. 
	
	Despite recent advance, there is still a huge gap between an Intelligence Agents and the human brain in several research areas that need reasoning about relations and graph-structured data such as scene graph \cite{battaglia2018relational} and natural language understanding. Humans can quickly recognize objects, their locations on grids, and Euclidean data such as images, infer their relationships, identify activities, and respond to random questions regarding the image. Modeling a system with computer vision and natural language capabilities that can answer random questions about an image seemed inspiring. 
	
	Solving the mentioned and related issues efficiently can give rise to a number of possible applications.  For example, visually challenged people can benefit from visual scene understanding, which allows them to obtain information about a scene via generated descriptions and ask questions about it. Understanding surveillance video is another use.\cite{baumann2008review}, autonomous driving\cite{kim2018textual}, visual commentator robot, human-computer interaction\cite{rickert2007integrating}, city navigation\cite{de2017guesswhat}, etc. Solving these issues frequently folds higher reasoning from an image content. Given the extensive span of fundamental and applied research, various surveys have been conducted in recent years to offer a thorough overview of vision and language task integration. On the other hand, these studies have focused on particular tasks involving the integration of language and vision, such as image
	description\cite{bernardi2016automatic,bai2018survey,hossain2019comprehensive}visual question answering\cite{kafle2017visual,wu2017visual},action recognition\cite{gella2017analysis} and visual semantics\cite{liu2019visual}.
	
	In this study, we provide a thorough and an extensive review of the state of the arts approaches, key models design principles and discuss existing datasets and methods for VQA and Visual reasoning task to understand vision and language representation learning. First, this paper introduce VQA and Visual reasoning as an example tasks of vision and language representation learning. Second, we explore existing annotated datasets driven tremendous progress in these fields in detail. Third, the paper further provide existing and state of the arts methods for VQA and Visual reasoning. Finally, we discuss some problems and possible future research directions.
	\section{Vision and Language}
	Vision-and-Language (V+L) research is a fascinating field at the intersection of CV and NLP, and it is gaining a lot of attention from both groups. A number of V+L challenges have spurred great progress in combined multimodal representation learning, which has been benchmarked over large-scale human-annotated datasets. The foundation of V+L is the visual understanding topic such are the popular ResNet, which extract CNN features. Secondly is the language understanding which is end goal is multimodal learning.
	\begin{figure}
		\centering
		\includegraphics[width=\linewidth]{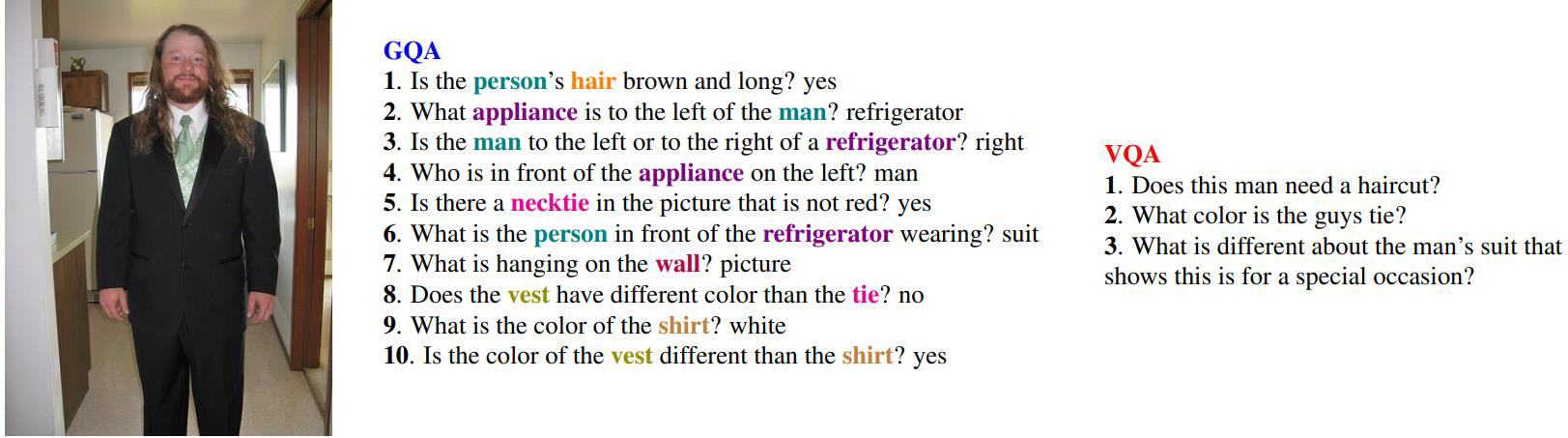}
		\caption{ Differences in VQA and visual reasoning dataset typical questions \cite{hudson2019gqa}.}
	\end{figure}
	\subsection{Vision and Language Integration Tasks}
	In recent years, substantial progress has been achieved in the investigation of the integration of language and vision. There are several activities that mix language observed at various levels\cite{zakari2021systematic}. Visual Question-Answering is an example of these types of multidisciplinary challenges, in which the model must comprehend the scene and objects depicted by an image, as well as their relationships, in order to respond to natural language queries about them. However, it was demonstrated in the first half of the year 2017 that these models perform poorly for complex questions that require understanding of a large number of object attributes and their relationships, prompting the community to look for models that perform more reasoning steps, giving rise to a new concept: Visual Reasoning.
	Several datasets, including CLEVR\cite{johnson2017clevr} and GQA\cite{hudson2019gqa}, have been developed to evaluate different reasoning skills, including spatial understanding and higher-level abilities like counting, logical inference, and comparison. These datasets frequently have tough compositional challenges, necessitating multi-step inferences. This differs from conventional VQA datasets, where the questions might be oversimplified and contain real-world bias. Figure 1 depicts the differences between the two tasks.  The goal of the visual reasoning challenge is to dissuade the model from building a cheating process by remembering statistical patterns or bias in the dataset; hence, in order to perform effectively, the model must understand the underlying reasoning process.
	
	\subsection{VQA and Visual Reasoning}
	The main idea behind Visual Question Answering (VQA) is to build a model that understands visual content in order to find relationships between pairs of natural language questions and answers. Images and videos are included in the visual information for VQA, however for the sake of this study, we will just look at images. Merging the visual representation acquired from a CNN and the text representation obtained from an embedding recurrent network is the most frequent technique for solving the VQA task. Humans, on the other hand, do not answer questions by studying the image and  question separately; instead, we focus on different areas of the image depending on the question
	
	\begin{figure}
		\centering
		\includegraphics[width=\linewidth]{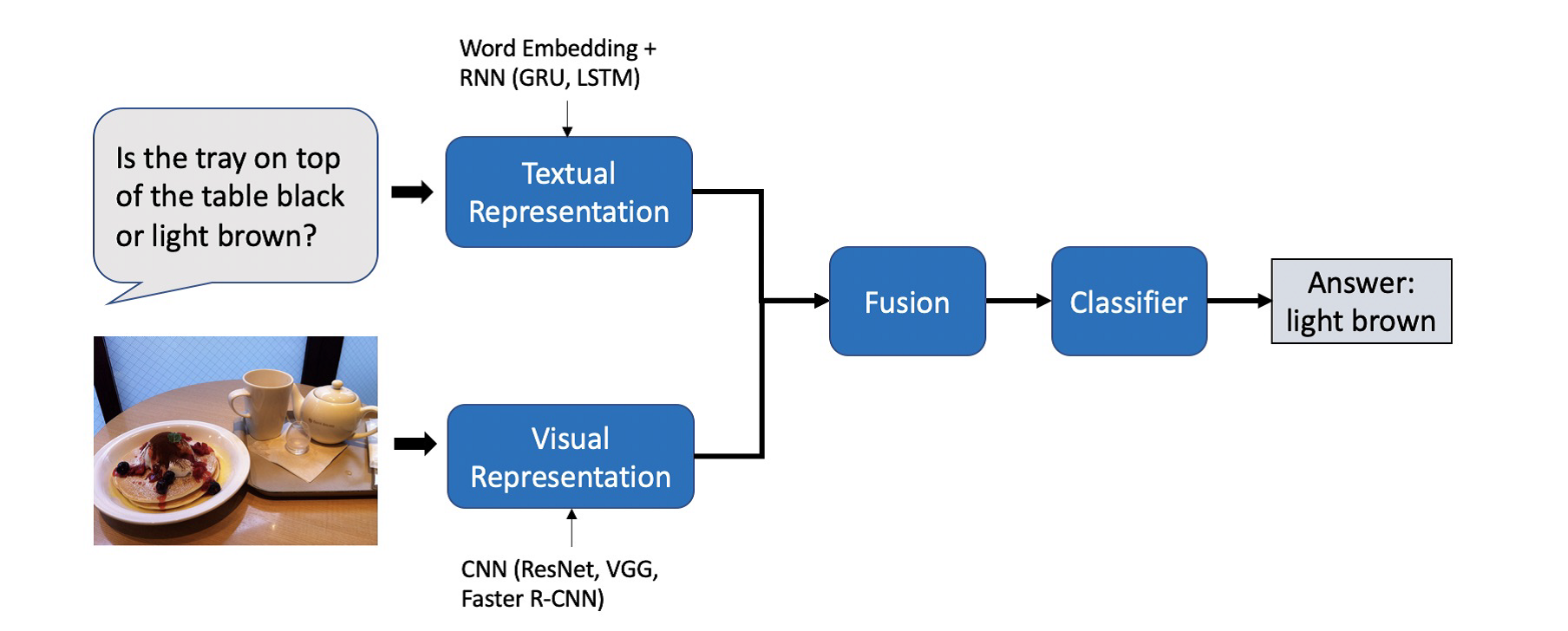}
		\caption{Typical VQA and Visual Reasoning framework}
		\label{fig:duck}
	\end{figure}
	
	\begin{table}[!t]
		\centering
		\resizebox{\textwidth}{!}{
			\begin{tabular}{lcccccccc} \toprule
				Datasets & Images & Questions & Question Type & Image Type & Task Focus & Venue&OE/MC\\
				\midrule
				DAQUAR\cite{malinowski2014multi} & 1,449 & 12,468 &4&Natural&VQA&NIPS 2014&OE   \\
				COCO-QA\cite{ren2015exploring} & 123,287& 117,684 &4&Natural&VQA&-&OE\\
				VAQ V1.0\cite{agrawal2017c} & 204k & 614K &-&Natural&VQA&ICCV 2015&OE\\
				VQA V2.0\cite{goyal2017making} & 204k & 1.1M &-&Natural&VQA&CVPR 2017&Both\\
				CVR\cite{zellers2019recognition} & 110k & 290K&&Natural&VR&CVPR 2019&MC\\ 
				GQA\cite{hudson2019gqa} & 113,018 &22,669,678&-&Natural&VR&CVPR 2019&OE\\ 
				RAVEN\cite{zhang2019raven} &1,120,000 &70,000 &4& Natural&VR&CVPR 2019&MC\\
				NLVR\cite{suhr2017corpus} &387,426&31,418&-&Synthetic&VR&ACL 2019&OE\\
				OK-VQA\cite{marino2019ok} &14,031&14,055&VQA&Natural&VQA&CVPR 2019&OE\\
				VizWiz\cite{gurari2018vizwiz} &-&31,173&-&Natural&-&CVPR 2018&OE\\
				KVQA\cite{shah2019kvqa} & 24K&&&Natural&-&AAAI 2019&OE\\ 
				CLEVR\cite{johnson2017clevr} &100,000&999,968&90&Natural&VR&CVPR 2017&OE\\ 
				FM-IQA\cite{gao2015you} &158,392&316,193&–&Natural& - &CVPR 2017&OE\\
				NLVR2\cite{suhr2019nlvr2} &107,292&29,680&-&Synthetic&VR&ACL 2019&OE\\
				TextVQA\cite{singh2019towards} &28,408&45,336&-&Natural&VQA&CVPR 2019&OE\\
				FVQA\cite{wang2017fvqa} &2190&5826&12&Natural&VR&CVPR 2019&OE\\
				VISUAL GENOME\cite{krishna2017visual} &108,000&145,322&7&Natural&VR&-&OE\\ 
				VQA-CP\cite{agrawal2018don} &&&&Natural&VQA&CVPR 2018\\
				Visual Madlibs\cite{yu2015visual} &10,738&360,001&12&Natural&VR&-&Fill-in-the blanks\\ 
				SHAPES\cite{andreas2015deep} &15,616&244&–&Synthetic&VR &-&Binary\\ 
				KB-VQA\cite{wang2015explicit} &700&2402&23&Natural& VQA&IJCAI 17&OE\\
				ICQA\cite{hosseinabad2021multiple} &42,021&260,840&-&Synthetic&VQA &-&OE\\
				DVQA\cite{kafle2018dvqa} &3,000,000&3,487,194&3&-&VQA &-&OE\\
				PathVQA\cite{he2020pathvqa} &4,998&32,795&7&-&VQA &-&MC\\
				Visual7w\cite{zhu2016cvpr} &47,300&327,939&7&Natural&VQA &CVPR 16&MC\\
				KRVQA\cite{cao2021knowledge} &32,910&157,201&6&Natural&VR &-&MC\\
				\bottomrule
			\end{tabular}
		}
		\caption{A Main characteristics of major VQA and Visual Reasoning datasets.}
	\end{table}
	\section{Datasets}
	We have many large scale annotated datasets which are driving tremendous progress in this fields. The VQA domain is so complicated that a suitable dataset should be large enough to represent the wide variety of options inside questions and visual material in real-world settings. Indeed in the past recent years there are lots of popular datasets to address the challenge of VQA and Visual reasoning. We discuss the datasets that are frequently used for this difficult task in the sections that follow.
	\subsection{DAQUAR}
	The first significantly large VQA dataset used as a benchmark was the dataset for Question Answering on Real-World Images (DAQUAR) \cite{malinowski2014multi}. The dataset, made accessible in 2015, was the first ever made for this task. Images were obtained from the NYU-Depth V2 dataset \cite{silberman2012indoor}. Every image in the collection has a label that describes what it is that it represents. With the aid of nine pre-made question templates, the question-answer combinations were automatically generated, and there are 894 different object classes from which the objects can be classified. Answers to these questions were given using data from the NYU-Depth V2 dataset\cite{silberman2012indoor}.
	\begin{table}[!t]
		\centering
		\resizebox{\textwidth}{!}{
			\begin{tabular}{lcccc} 
				\hline Split & Images & Images with captions & Total Captions & Categories \\
				\hline Train & 113,287 & 5 & 566,435 & $-$ \\
				Val & 5,000 & 5 & 25,000 & $-$ \\
				Test & 5,000 & 5 & 25,000 & $-$ \\
				\hline Total & 123,287 & 5 & 616,435 & 80 \\
				\hline
			\end{tabular}
		}
		\caption{MSCOCO image dataset Splits.}
	\end{table}
	
	Despite being a groundbreaking dataset for VQA, the DAQUAR is too small to efficiently train and test more complex models. The amount of questions is limited by DAQUAR's small size and the fact that it only features indoor scenes. The images often have noticeable clutter and, in some cases, extreme lighting. As a result, many questions are extremely difficult to answer, and even humans often struggle to select the correct answer with approximately 51\% in accuracy.
	\subsection{COCO-QA}
	
	The COCO-QA dataset\cite{ren2015exploring}, uses images from the MS-COCO dataset\cite{lin2014microsoft} with human-annotated captions. The dataset has 123 287 images, of which 78 736 are used for training and 38 948 for testing QA pairs. The accuracy of the predicted answer is measured using the WUPS score, which is very similar to the DAQUAR\cite{malinowski2014multi} dataset. There is one question and one answer for each image in the datasets. The automatically generated questions have a high recurrence rate, even though the datasets increases the training data for VQA tasks.
	
	The major drawback of this datasets is due to the weakness in the NLP algorithm that was used to produce the QA pairs. In order to make processing longer sentences easier, they are often divided into smaller chunks, but in many of these instances, the algorithm struggles to deal with the appearance of clauses and linguistic differences in the structure of the sentence. This leads to poorly phrased questions, resulting in many of which have misspellings errors, and some of which are completely incomprehensible.
	The fact that there are only four different types of questions, all of which are constrained to the aspects discussed in COCO's captions, is another significant drawback.
	
	\subsection{Real-World Images}
	\subsubsection{VQA v1.0}
	One of the most popular VQA datasets is VQA-v1.0 \cite{antol2015vqa}, which was built using the COCO dataset\cite{ren2015exploring} and had two parts: VQA-v1-real, which uses real photos, and VQA-v1.0-abstract, which uses artificial cartoon images. From the COCO dataset, VQA-v1-real uses 123,287 pictures for training and 81,434 for testing. The QA pairs are gathered by human annotators, resulting in great diversity, and binary (for example, yes/no) questions are included. Each of the 614,163 questions has 10 responses from 10 distinct annotators, totaling 614,163 questions. However, this dataset has a significant bias, making it possible to provide answers to several queries without any prior visual knowledge\cite{malinowski2014multi}.
	\begin{table}[!t]
		\centering
		\resizebox{\textwidth}{!}{
			\begin{tabular}{lccc|cc} 
				\hline Dataset & Real & Questions & Answers & \multicolumn{2}{|c}{ Textual Annotations } \\
				Split & Scenes & per Image & per Question & Questions & Answers \\
				\hline Training & 82,783 & 3 & 10 & 248,349 & $2,483,490$ \\
				Validation & 40,504 & 3 & 10 & 121,512 & $1,215,120$ \\
				Test & 81,434 & 3 & 10 & 244,302 & $2,443,020$ \\
				\hline
			\end{tabular}
		}
		\caption{VQA v1.0 dataset with real scenes Splits}
	\end{table}
	
	The biggest concern with the VQA dataset was that of bias inherent. Lingual priors had a very huge impact on the answers to the questions in the datasets, despite the authors of \cite{antol2015vqa} showing that the models prove to be effective in answering the questions when given the image. The concept of the revised version 2.0 of the VQA dataset was inspired by this bias in the dataset and the huge impact of the lingual priors on the answers.
	\subsubsection{VQA v2.0}
	VQA v2.0 \cite{goyal2017making} is the updated edition of the VQA dataset.443757 image pairs make up the training set, 214354 image pairs make up the validation set, and 447793 image pairs make up the test set [9]. Its dataset is two times as large as that of the second edition. The data set contains 1.1 million (image, question) pairs and 13 million connected answers from VQA v2.0, all arbitrarily annotated. Furthermore, for each question in the information, there are two comparable images \cite{zhang2016yin}, but the answers are distinct, resulting in a more equal statistical distribution of the image database. VQA v2.0 datasets are significantly larger and more centered real-world images from MS COCO than VQA v1.0 datasets\cite{lin2014microsoft}.
	
	The balanced VQA v2.0 dataset reduces the biases present in the original VQA v1.0 real dataset, preventing VQA models from taking advantage of linguistic priors to increase evaluated scores and assisting in the development of highly explainable VQA models that place a greater emphasis on the visual contents.
	\begin{table}[!t]
		\centering
		\resizebox{\textwidth}{!}{
			\begin{tabular}{lcc|ccc} 
				\hline Dataset & Real & Answers & \multicolumn{3}{|c}{ Textual Annotations } \\
				Split & Images & per Question & Questions & Answers & Complementary Pairs \\
				\hline Training & 82,783 & 10 & 443,757 & $4,437,570$ & 200,394 \\
				Validation & 40,504 & 10 & 214,354 & $2,143,540$ & 95,144 \\
				Test & 81,434 & 10 & 447,793 & $4,477,930$ & $-$ \\
				\hline
			\end{tabular}
		}
		\caption{Splits of the VQA v2.0 dataset with real images.}
	\end{table}
	
	\subsubsection{Visual Genome}
	Genome Visualization\cite{krishna2017visual} the goal of Visual Genome is to improve performance on cognitive activities, particularly spatial connection thinking. Over 108,000 photos are included in the dataset, with an average of 35 items, 26 characteristics, and 21 pairwise interactions between objects. It highlights the importance of relationships and attributes in the annotation region, even though they are critical for visual understanding. As a result, it accumulates around 50 descriptions for distinct image elements in this dataset.The key main advantage of the Visual Genome dataset for VQA is the ability to use organized scene annotations. 
	
	\begin{figure}
		\centering
		\includegraphics[width=\linewidth]{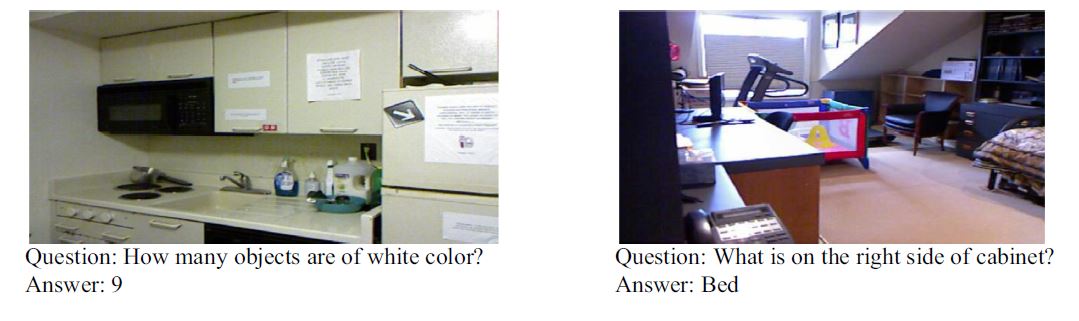}
		\caption{Sample of images images from the DAQUAR dataset \cite{malinowski2014multi} }
	\end{figure}
	\begin{figure}
		\centering
		\includegraphics[width=\linewidth]{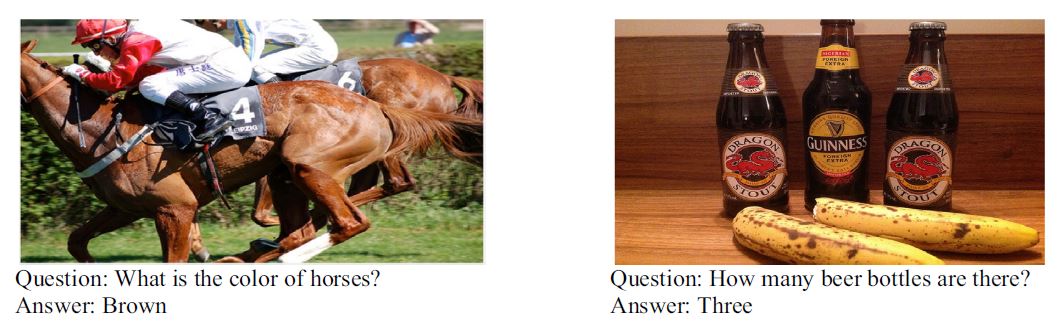}
		\caption{Sample of images from the COCO-QA dataset \cite{ren2015exploring} }
	\end{figure}
	\begin{figure}
		\centering
		\includegraphics[width=\linewidth]{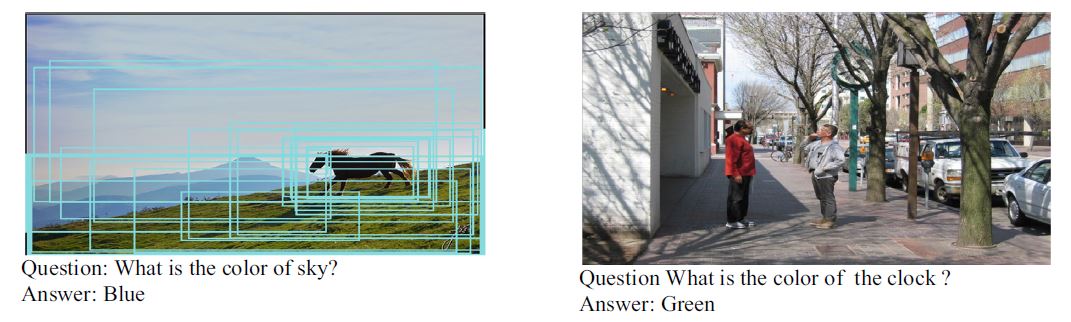}
		\caption{Sample of images from the Visual Genome dataset \cite{krishna2017visual} }
	\end{figure}
	\begin{figure}
		\centering
		\includegraphics[width=\linewidth]{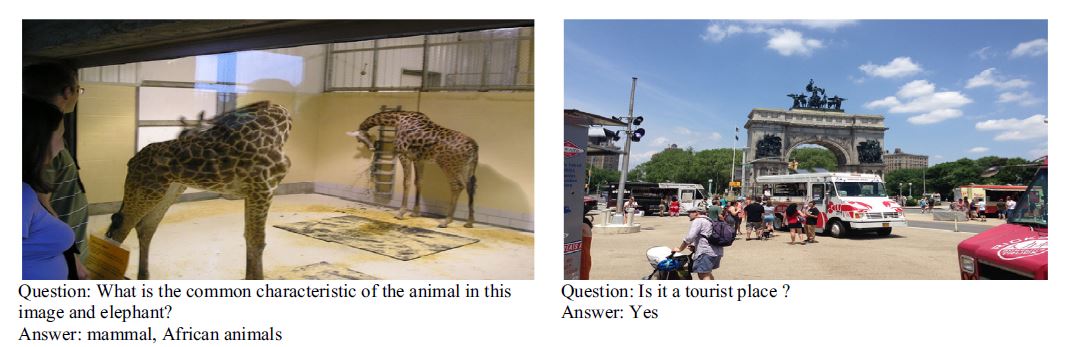}
		\caption{Knowledge bases images from datasets KB-VQA \cite{wang2015explicit}}
	\end{figure}
	\begin{figure}
		\centering
		\includegraphics[width=\linewidth]{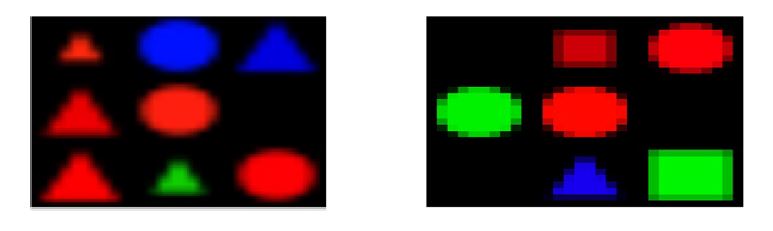}
		\caption{Shapes \cite{szegedy2015going} dataset image. There are three types of questions: counting, spatial reasoning and inference.}
	\end{figure}
	
	\subsubsection{Visual 7W}
	The Visual 7W dataset \cite{zhu2016visual7w} is a subset of the Visual Genome dataset \cite{krishna2017visual}, which contains 327,939 QA pairs and 47,300 pictures from MS COCO\cite{ren2015exploring}. It contains 1,311,756 multiple-choice questions created by human annotators, with one correct answer and three wrong answers for each question. What, where, when, who, why, how, and which are among the seven categories (7W) of questions gathered. In Visual 7W, the average length of a response is 2.0 words, and 27.6\% of question-answer combinations contain a lengthier answer, meaning more than two words\cite{zhu2016cvpr}. Simple yes/no questions with binary responses are not allowed.
	
	\subsubsection{GQA}
	GQA is one of Stanford's most recent datasets \cite{hudson2019gqa}, and it focuses on scene comprehension and reasoning. It contains 113,018 real-world pictures from Visual Genome, as well as their related scene graphs. The objects, characteristics, and relations inside scene graphs are extensively normalized and rectified in order to acquire more exact annotations and create high-quality queries. In addition, the GQA dataset contains 22,669,678 multi-step questions created by a question engine with rich material obtained from scene graphs, as well as 524 structural patterns, 250 of which are manually constructed and 274 retrieved from the VQA 1.0 dataset.
	\begin{table}[!t]
		\centering
		\resizebox{\textwidth}{!}{
			\begin{tabular}{ccc|cccc} 
				\hline Images & Questions & Vocabulary & Train (\%) & Validation (\%) & Test (\%) & Challenge (\%) \\
				\hline 113,018 & $22,669,678$ & 3,097 & $70 $ & $10 $ & $10 $ & $10 $ \\
				\hline
			\end{tabular}
		}
		\caption{GQA dataset characteristics and splits.}
	\end{table}
	Its semi-synthetic nature is also the cause of several drawbacks. This is aa a result of the questions being synthetic in nature, and they have insufficient linguistic diversity. The template-based creation can also result in odd wording. Additionally, producing a dataset of this size encourages the development of noisy annotations. The occurrence of ambiguous questions is fairly common.
	\subsubsection{VCR}
	VCR\cite{zellers2019recognition} is a relatively large dataset that was developed for visual reasoning at the cognitive level. There are approximately 110k photos, 290k multiple choice questions, and 290k correct answers and rationales in all. It is quite diversified, making it difficult to work with. Table 6 shows the dataset's official divides as well as some high-level statistics.
	
	VCR is considered a multiple-choice QA. In contrast to conventional VQA, it is essential to presume the correct answer with commonsense reasoning. In addition, VCR requires that the model consider the correct answer and its supporting evidence in addition to giving the correct answer. It is crucial to remember that questions, answers, and reasons are composed of both textual and visual words; the latter refers to related object tags that link the textual description and the referred image regions.
	\begin{table}[!t]
		\centering
		{
			\begin{tabular}{l|ccc}
				\hline Dataset Characteristic & Train & Validation & Test \\
				\hline No. of questions & 212,923 & 26,534 & 25,263 \\
				No. of answers per question & 4 & 4 & 4 \\
				No. of rationales per question & 4 & 4 & 4 \\
				\hline No. of images & 80,418 & 9,929 & 9,557 \\
				No. of movies  & 1,945 & 244 & 189 \\
				\hline Avg. question length & $6.61$ & $6.63$ & $6.58$ \\
				Avg. answer length & $7.54$ & $7.65$ & $7.55$ \\
				Avg. rationale length & $16.16$ & $16.19$ & $16.07$ \\
				Avg. num. of objects  & $1.84$ & $1.85$ & $1.82$ \\
				\hline
			\end{tabular}
		}
		\caption{High-level statistics of the VCR dataset.}
	\end{table}
	
	\subsubsection{VizWiz}
	VizWiz \cite{gurari2018vizwiz} is the first widely accessible vision dataset coming from people that can not see (blind).Additionally, since it is the first VQA dataset, the challenge of predicting whether or not a visual question can be answered is naturally encouraged. It is based on previous work [8] that collected 72,205 visual questions over the course of four years using the VizWiz app, which is accessible on both iPhone and Android phones.
	
	The perception side of VizWiz defines as one of the major hurdles because it must deal with low-quality images. The evaluation of reasoning is therefore given less weight. However, it still calls for various fascinating abilities, including the ability to tell when a question can be answered, reading, counting, and decode evasive questions, among others.
	\subsubsection{RAVEN}
	The RAVEN dataset \cite{zhang2019raven} was created to test visual thinking in relational and analogical contexts. Raven's Progressive Matrices (RPM) are used to construct it \cite{burke1958raven}. In a hierarchical representation, it also links vision to structural, relational, and analogical reasoning. The dataset is divided into three sections: training, validation, and testing, in the order of 6:2:2. The dataset's statistics are presented in Table below.
	\begin{table}[!h]
		\begin{center}
			\begin{tabular}{lr}
				\hline Number of images & 1,120,000 \\
				RPM & 70,000 \\
				Tree-structure & 16 \\
				Labels & 1,120,000 \\
				Annotations & $440, 000$ \\
				Avg. rules & $6.29$ \\
				\hline
			\end{tabular}
			\caption{characteristics of the RAVEN dataset}
		\end{center}
	\end{table}
	\subsubsection{VQA-CP}	
	The VQA-CP\cite{agrawal2018don}was created in order to reduce the influence of statistical bias in answers by altering the distribution of answers for each type of question in the training and test sets. The data in the VQA v1\cite{antol2015vqa}and VQA v2\cite{goyal2017making} were divided and restructured using question grouping and greedy re-splitting approaches, respectively. When questions are grouped, several questions that are prepared for the same purpose and have the same real-world solution are produced. A greedy technique was then developed to split the data into training and test sets, taking into account how to maximize the number of ideas and prevent having groups repeat the training and test sets. The training set has 118,500 images, 245,500 questions, and 2.5 million answers, whereas the test set has 87,500 images, 125,500 questions, and 1.3 million answers.
	
	The VQA-CP dataset provides a challenging environment for testing VQA models. VQA models cannot take advantage of language bias because the train and test sets have different semantics. The same object, however, can be present in both the train and test sets because the train and test separations are only implemented based on question types and answers. For instance, even though the train and test sets are divided by various question types, the image of a cat can appear in both. In addition, there are fewer test questions than in other benchmark VQA datasets because the questions are repurposed from the train-Val split of the VQA datasets. Despite this, the VQA-CP dataset provides a solid evaluation framework to test a VQA model's performance independence from linguistic bias.
	\subsection{Synthetic Images}
	\subsubsection{CLEVR}
	Compositional Language and Elementary Visual Reasoning(CLEVR)\cite{johnson2017clevr} is a diagnostics dataset that has 100,000 images and 864,968 questions, with the goal of examining explicit visual reasoning skills in VQA. As ground-truth annotations of photographs, the CLEVR collection offers item properties such as size, shape, material, and color, as well as spatial coordinates. In CLEVR, questions are generated automatically based on 90 question families, each of which has one program template and four text templates on average. Each question is linked to a functional program that includes a set of basic operations including object counting, attribute querying, and comparing.         
	
	This dataset is limited in its ability to represent a wide range of possible real-world scenarios and users. For instance, lighting, object diversity, and image ambiguity which are the three main issues for real-world images are not always reflected in computer-generated images. Equally, the questions do not portray the diversity of visual questions that would be exciting to actual VQA system users.                                
	\begin{table}[!t]
		\centering
		\resizebox{\textwidth}{!}{
			\begin{tabular}{l|cccc}
				\hline Split & Images & Questions & Unique Questions & Overlap with train \\
				\hline Train & 70,000 & 699,989 & 608,607 & $-$ \\
				Val & 15,000 & 149,991 & 140,448 & 17,338 \\
				Test & 15,000 & 149,988 & 140,352 & 17,335 \\
				\hline Total & 100,000 & 999,968 & 853,554 & $-$ \\
				\hline
			\end{tabular}
		}
		\caption{CLEVR dataset Splits.}
	\end{table}
	\subsubsection{SHAPES}
	Berkeley created SHAPES \cite{andreas2016neural}, a dataset made up of synthetic pictures with 2D abstract shapes that may be used to study explicit reasoning in visual question responding. This dataset contains 15,616 synthetic pictures and 244 yes-no binary questions. SHAPES objects are very basic, with only a few fundamental characteristics and spatial relationships. Furthermore, due to the tiny size of question number, the diversity of questions is limited. Each image is a 30×30 RGB image depicting a 3×3 grid of objects. Each object is characterized by shape (circle, square, and triangle), color (red, green, and blue) and size (small, big).
	
	Although using real-world imagery cannot be replaced by SHAPES, the concept behind it is very helpful. An algorithm may be limited in its ability to analyze images if it struggles to perform well on SHAPES but excels on other VQA datasets.
	\subsubsection{NLVR}
	NLVR \cite{suhr2017corpus} is a multimodal dataset which includes human language words based on synthetic visuals. Different things, such as triangles, circles, and squares, are encapsulated in the photos. These items come in a variety of sizes and are positioned in various locations inside the image. Crowd workers wrote the descriptions for the images by hand. Figure 9 shows the splits of the datasets. 
	\begin{table}[!t]
		\centering{
			\begin{tabular}{l|cc}
				\hline Split & Unique Sentences & Examples \\
				\hline Training & 3,163 & 74,460 \\
				Validation & 267 & 5,940 \\
				Test-Public & 266 & 5,934 \\
				Test-Unreleased & 266 & 5,910 \\
				\hline Total & 3,962 & 92,244 \\
				\hline
			\end{tabular}
		}
		\caption{NLVR dataset Splits.}
	\end{table}
	\subsubsection{NLVR2}
	NLVR2\cite{suhr2018corpus} dataset was developed to be resilient to language bias. Because of the synthetic nature of the NLVR dataset, issues such as restricted expressivity and semantic diversity have been addressed in the next generation of NLVR, dubbed Natural Language for Visual Reasoning. NLVR2 includes a pair of visuals as well as a grounded natural language description, similar to NLVR.
	\begin{table}[!t]
		\centering{
			\begin{tabular}{l|cc} 
				\hline Split & Unique Sentences & Examples \\
				\hline Training & 23,671 & 86,373 \\
				Validation & 2,018 & 6,982 \\
				Test-Public & 1,995 & 6,967 \\
				Test-Unreleased & 1,996 & 6,970 \\
				\hline Total & 29,680 & 107,292 \\
				\hline
			\end{tabular}
		}
		\caption{NLVR2 dataset Splits.}
	\end{table}
	\subsubsection{VQA-abstract}
	Abstract sceneries, as previously indicated, are a part of the VQA benchmark \cite{ren2016faster}. VQA-abstract includes 50,000 "realistic" abstract sceneries to enhance real-word visuals and encourage researchers to focus on higher-order thinking rather than fundamental vision tasks. Indoor and outdoor scenes with various items, animals, and humans are collected, and each abstract scene has five captions to define the image's content.
	In the same way as VQA-real, 150,000 question-answer pairs are generated for VQA-abstract, and the pairs are divided into 60,000 for training, 30,000 for validation, and 60,000 for test. The open-answer question type with ten plausible options and the multi-choice assignment with one correct answer and 17 incorrect candidates are both available in VQA-real. Furthermore, both genuine photos and abstract scenarios had comparable distributions in terms of duration and sorts of queries.
	
	\subsection{External Knowledge Images}
	\subsubsection{KB-VQA}
	The KB-VQA dataset \cite{wang2015explicit} was the first to incorporate an external knowledge base. It was built with the goal of evaluating VQA algorithms' performance on questions that demand higher-level knowledge and explicit reasoning about image contents using external data. It includes 700 photos that were chosen from MS-COCO. The photos are chosen to cover approximately 150 object classes and 100 scene classes, with each image typically displaying 6 to 7 things. There are 2,402 questions in total, all of which are prepared by human annotators using 23 templates. As an external knowledge base, DBpedia \cite{auer2007dbpedia} is used.
	\subsubsection{KVQA}
	The KVQA (Knowledge-aware VQA) dataset \cite{shah2019kvqa} was created with the goal of emphasizing questions that require external knowledge. There are 183K question-answer pairs in the collection, with around 18K people enclosed within 24K photos. To answer the questions in this dataset, you'll need to use multi-entity, multi-relation, and multi-hop reasoning over KG. Another unique feature of our dataset is inquiries that go beyond KG entities as ground-truth answer.
	
	\begin{table}[!h]
		\begin{center}
			\begin{tabular}{lr}
				\hline No. of images & 24,602 \\
				No. of QA pairs & 183,007 \\
				No. of unique entities & 18,880 \\
				No. of unique answers & 19,571 \\
				Avg. question length & $10.14$ \\
				Avg. answer length & $1.64$ \\
				Avg. number of questions per image & $7.44$ \\
				\hline
			\end{tabular}
			\caption{characteristics of the KVQA dataset}
		\end{center}
	\end{table}
	
	\begin{table}[!t]
		\centering{
			\begin{tabular}{lccc}
				\hline Split & Percent (\%) & Images & Q\&A pairs \\
				\hline Train & 70 & $17,000$ & $130,000$ \\
				Validation & 20 & $5,000$ & $34,000$ \\
				Test & 10 & $2,000$ & $19,000$ \\
				\hline
			\end{tabular}
		}
		\caption{KVQA dataset Splits}
	\end{table}
	
	KVQA contains challenges for both the vision community, e.g., handling faces with different poses, race and scale, and large coverage of persons, as well as for the language understanding community, e.g., answering questions that require multi-hop and quantitative reasoning over multiple entities in Knowledge Graph. furthermore, biased and unbalanced answers in VQA datasets have been a challenge, and often blindly answering without understanding image content leads to superficially high performance. further, the biased and unbalanced answers in VQA
	datasets have been a challenge, and often blindly answering
	without understanding image content leads to a superficially
	high performance.
	\subsubsection{FVQA}
	Fact-based VQA (short for FVQA) \cite{wang2017fvqa} is a follow-up to KB-VQA that incorporates more organized knowledge bases like ConceptNet \cite{liu2004conceptnet} and WebChild \cite{tandon2014acquiring}. The MS-COCO and ImageNet databases yielded a total of 2,190 photos, with 326 object classes, 221 scene classes, and 24 action classes among them. There are 5,826 questions (equivalent to 4,216 distinct facts) that 38 people are collecting together. Longer questions are available in the FVQA dataset, with an average length of 9.5 words.
	\subsubsection{OK-VQA}
	OK-VQA \cite{marino2019ok} is based on a subdivision of MSCOCO\cite{ren2015exploring},Currently this is the largest knowledge-based VQA dataset available and includes annotations such as questions, answers, knowledge categories, and so on.This dataset outperformed earlier research in terms of scale, question quality, and image quality. Additionally, it possesses the trait of not being "closed" or explicitly derived from a specific source, and could be referred to as "open"-domain knowledge. The dataset includes 14,031 images and 14,055 questions on a range of subjects, such as travel, materials, sports, cooking, geography, plants, animals, science, the weather, and more.Table 13 illustrates the dataset's statistics. 
	\begin{table}[!t]
		\centering{
			\begin{tabular}{lr}
				\hline No. of images & 14,031 \\
				No. of Questions & 14,055 \\
				Answer per Question & 5 \\
				No. of unique entities & 12,591 \\
				No. of unique answers & 14,454 \\
				Unique question length (words) & $7,178$ \\
				Total categories & $10+1$ \\
				Average Answer length (words) & $1.3$ \\
				\hline
			\end{tabular}
		}
		\caption{characteristics of the OK-VQA dataset.}
	\end{table}
	\subsection{Others}
	\subsubsection{Diagrams}
	AI2 Diagrams is a dataset proposed by \cite{kembhavi2016diagram} for VQA on diagrams (AI2D). It has almost 5,000 diagrams that depict elementary school science topics including the water cycle and the digestive system. Segments and relationships between graphical elements are annotated on each diagram. There are about 15,000 MC questions and answers in the collection.The authors suggest a technique in the same paper to infer accurate results from this dataset. The approach creates systematic representations known as diagram parse graphs (DPG) using methods designed especially for diagrams, such as OCR for text recognition or arrow recognition. The correct answers are then deduced using the DPGs. The visual parser of diagrams is still tricky compared to VQA on natural images, and the questions frequently demand advanced levels of reasoning, making the task, in general, very difficult.
	\section{Evaluation Metrics }
	Evaluating a natural language question that is developed automatically by a VQA or Visual reasoning framework is a difficult task. It is required to consider either syntactic (grammatical) and semantic validity. A question in VQA or Visual reasoning task can be open-ended, requiring the system to establish a response. Phrase to answer the question or a MC in which the framework chooses one of the options provided.
	\begin{equation}
		\text { Accuracy }=\frac{\text { questions answered correctly }}{ \text { total questions }}
	\end{equation}
	Very simple accuracy may always be used to measure the multiple-choice VQA task when a criteria obtains the appropriate reply, regardless of whether it decides the correct alternative. When an algorithm's projected reply properly has the actual ground truth, very simple accuracy may in addition be used to gauge open-ended VQA and even VR tasks. Typically, the drawback of this specific basic accuracy figure is that it demands an actual fit. Consider the next visual question: 'What fruits are provided inside the image?" In the event that the algorithm comes back with "orange," although the right content label is "oranges," it is viewed as erroneous; similarly, in the event that the system comes back with "banana," it is considered completely wrong.
	Wu-Palmer The second evaluation metric (WUPS) \cite{wu1994verb}, that was developed as an alternative to basic accuracy is similarity. This measure attempts to quantify the semantic connection difference between an algorithmically predicted answer and available ground truth in the datasets. Based on the proximity of ground truth answers in datasets and anticipated by the algorithm to a question, WUPS will award a number between 0 and 1. For example, the similarity score between orange and oranges is 0.88, whereas the similarity score between apple and fruit is 0.76. WUP(a,b) will return the position of phrases 'a' and 'b' in relation to the value of the Lowest Common Dataset obtained based on the taxonomy tree, with NQ denoting the total number of questions, PA denoting the set of predicted answers, and GA denoting the sequence of ground truth answers (a, b).
	
	\begin{equation}
		\operatorname{WUPS}(a, b)=\frac{1}{N_{Q}} \sum_{i=1}^{N_{Q}} \min \left\{\prod_{a \in P_{A}} \max _{t \in G_{A}} W U P(a, t), \prod_{t \in G_{T}} \max _{a \in P_{A}} W U(a, t),\right\}
	\end{equation}
	The total number of questions is $N_{Q}$, the set of predicted answers is $P_{A}$, the set of ground truth answer is GA, and the set of ground truth answers is WUP(a,b). Based on the taxonomy tree, it will output the position of terms 'a' and 'b' in respect to the location of the Lowest Frequent Dataset obtained (a, b).
	Obtaining numerous independent ground-truth answers for each question is another way to evaluate the VQA and Visual Reasoning task. This is known as the consensus metric . The VQA dataset\cite{antol2015vqa} was created with it, f For each question in the VQA dataset, ten distinct subjects gathered the ground-truth answers. On the VQA dataset\cite{antol2015vqa}, this evaluation is carried out by comparing a computed answer to the ten ground truth answers supplied by ten different subjects as provided by:
	\begin{equation}
		\text { Accuracy }_{V QA }=\min \left(\frac{\text { A precise answer is given by } \# \text { subjects }}{3}, 1\right)
	\end{equation}
	An answer is regarded 100\% accurate if it is provided by at least three subjects according to the above equation. This statistic has its own set of limitations. For starters, some questions may have two correct answers. Second, gathering ground truth answers for each question is prohibitively expensive. Third, inter-human consensus is poor for 'Why' type of questions because it is extremely difficult for three subjects to give exactly the same answer.
	\section{Methods}
	In the last five years, a great number of VQA and visual reasoning methods have been developed. All known approaches, on the other hand, consist on extracting features from the question and the image, then combining the features to give an answer. Bag-Of-Words (BOW) \cite{qader2019overview}, Long Short Term Memory (LSTM)\cite{staudemeyer1909understanding}, gated recurrent units (GRU), encoders, and skip-thought vectors can all be utilized for text. CNNs pre-trained on ImageNet are the most popular choice for image features. When it comes to generating answers, most techniques represent the problem as a classification exercise. As a result, the primary difference between these techniques is how they merge textual and visual data. For instance, concatenate and run them through a linear classifier. Alternatively,  Bayesian models may be used to indicate the core relationships between the distributions of question, image, and answer features.
	In this section we go through several recent architectures that have been proposed for the VQA and Visual reasoning task. We categories these models into three main sections: External Knowledge, Neural Network and Explicit reasoning. In the following, we discuss each section in more details.
	\begin{table}[htbp]
		\centering
		\caption{Approach, Method and Models Summary}
		\resizebox{\textwidth}{!}{
			\begin{threeparttable}
				\begin{tabular}{lcccccc}
					\toprule
					Approach&   Methods&  Models&   Year& Image Futures & Question Encoding \\
					\midrule
					\multirow{20}{*}{Neural Network}& \multirow{3}{*}{Traditional neural network}
					& MDFNet  \cite{zhang2021multimodal}       &2020           & ResNet   &GRU\\  
					&   &iBOWIMG  \cite{zhou2015simple}     &2015           & GoogLeNet   &LSTM \\ 
					&   &VIS+LSTM  \cite{ren2015exploring}     &2015           &VGGNet   &LSTM \\
					&   &mQA  \cite{gao2015you}     &2015           &GoogleNet   &LSTM \\
					&  &DPPnet  \cite{noh2016image}       &2016           & VGGNet,    &LSTM, GRU\\
					&  &N-KBSN  \cite{ma2021joint}       &2021           & Faster R-CNN,    &ELMo \\
					
					\cmidrule {2 -6}
					&\multirow{6}{*}{Attention Mechanism}
					& SAN \cite{yang2016stacked}&2016 &VGGNet +GoogleNet&LSTM, CNN\\
					&   & MAC \cite{hudson2018compositional}&2018&ImageNet&biLSTM \\
					&   & BUTD  \cite{anderson2018bottom}&2018 &Faster R-CNN &LSTM \\ 
					&   & NS-CL \cite{mao2019neuro}   &2019  &ResNet    & - \\
					&   &SMem-VQA \cite{xu2016ask}   &2015  &GoogleNet    &CBOW, LSTM \\ 
					&   & HieCoAtt \cite{lu2016hierarchical} &2016 &VGGNet, ResNet &LSTM  \\
					&   & SelRes \cite{hong2020selective}   &2020       &Faster
					R-CNN    &GloVe \\
					&   &QLOB \cite{vu2020question}   &2020      &ResNet    &LSTM, GRU \\
					&   &ALSAs \cite{liu2020alsa}   &2020      &ResNet, Faster
					R-CNN    &GloVe, GRU \\
					\cmidrule{2-6}
					&\multirow{8}{*}{Bilinear pooling } 
					& MCB  \cite{fukui2016multimodal}   &2016     &VGGNet+ResNet    &LSTM, GRU \\
					&   & BAN \cite{kim2018bilinear}   &2018            &Faster R-CNN    &51.06 \\
					&   & MFB  \cite{yu2017multi}   &2017            &ResNet    &51.06 \\
					&   & MFH \cite{yu2018beyond}   &2018            &ResNet   &LSTM \\
					&   & Mutan \cite{ben2017mutan}   &2017            &ResNet    &GRU \\
					&   & Block \cite{ben2019block}    &2019            & Faster-RCN   &GRU \\
					&   &MLB  \cite{kim2016hadamard}    &2016            &ResNet    &GRU \\
					&   &QC-MLB \cite{vu2020question}   &2020      &ResNet    &Skip-thoughts \\
					\cmidrule{2-6}
					&\multirow{3}{*}{Transformer} & MCAN \cite{yu2019deep}   &2019        &Faster
					R-CNN    &GloVe+
					LSTM \\
					&   & Pixel-BERT \cite{huang2020pixel}   &2020       &Faster R-CNN    & BERT \\
					&   & SANMT \cite{zhong2020self}   &2020   &ResNet    &Skip-thoughts \\ 
					\cmidrule {1 -6}
					\multirow{9}{*}{Explict Reasoning}
					&\multirow{9}{*}{Modular Neural}&NMN \cite{andreas2015deep} &2016 &VGGNet  &LSTM\\
					&   &N2NMN \cite{hu2017learning} &2017            &VGG-16 &LSTM\\
					&   &FiLM \cite{perez2018film} &2018   &GRU &LSTM\\
					&   &Stack-NMN \cite{hu2018explainable} &2018            &CNN &BiLSTM\\
					&   &DMN \cite{xiong2016dynamic} &2016  &VGGNet &GRU\\
					&   &ReasonNet \cite{ilievski2017multimodal} &2017            & - &LSTM\\
					\cmidrule {1 -6}
					\multirow{4}{*}{External Knowledge}
					&\multirow{4}{*}{-}&FVQA \cite{wang2017fvqa} &2017         &VGG-Net &LSTM\\
					&   &Mucko  \cite{zhu2020mucko} &2020         &Faster				R-CNN &GloVe\\
					&   &Ahab  \cite{wang2015explicit} &2015        &Fast-RCNN&GloVe\\
					&   &KVL-BERT  \cite{song2021kvl} &2020         &Faster
					R-CNN &BERT\\
					&   &GRUC  \cite{yu2020cross} &2020         &Faster
					R-CNN &GloVe, LSTM\\
					&   &KRVQR  \cite{cao2021knowledge} &2020         &Faster
					R-CNN &GloVe, LSTM\\
					&   &KRISP  \cite{marino2021krisp} &2021         &Faster				R-CNN &BERT\\
					&   &KBSN  \cite{dey2021external} &2021         &Faster				R-CNN &ELMo\\
					
					\midrule
					\multirow{10}{*}{Others}& \multirow{2}{*}{Diagnosis Method}   
					& GVQA  \cite{agrawal2018don}       &2018           &VGG-Net   &LSTM \\ 
					&   & ViQAR  \cite{dua2021beyond}     &2021           &BERT     &LSTM \\
					\cmidrule {2 -6}
					&\multirow{2}{*}{Memory-Based}
					& VKMN \cite{su2018learning}&2018 &ImageNet   &LSTM \\
					&   & CM-VQA \cite{jiang2015compositional}&2015&GoogleNet&LSTM \\
					
					\cmidrule{2-6}
					&\multirow{8}{*}{Graph and Neural-Symbolic} 
					& LRTA  \cite{liang2020lrta}   &2020     &Scene Graph Generation    &Semantic Parsing \\
					&   & BGNVQA \cite{guo2021bilinear}   &2020            &Faster R-CNN    & - \\
					&   & ODA-GCN \cite{zhu2020object}   &2020            &GloVe    &GRU \\
					&   & LG-Capsule \cite{cao2020linguistically}   &2020            &ResNet   &GRU \\
					&   & ReGAT \cite{li2019relation}   &2019            &Faster-R-CNN    &RNN \\
					&   & MN-GMN \cite{khademi2020multimodal}    &2020            &Faster R-CNN    &GRU \\
					\bottomrule
				\end{tabular}
			\end{threeparttable}
			\label{tab:test}}
	\end{table}
	
	\subsection{Method based on External knowledge}
	Recently, there are growing interests in answering questions which require commonsense knowledge involving common concepts present in the image. Some VQA and visual reasoning task questions call for in-depth analysis that relies less on the visual information in the input image and more on general knowledge of human nature. For instance, the question "Which city is pictured here?" with an image of the Eiffel Tower implies that the Eiffel Tower is in Paris. Regular joint embedding methods have a limited ability to handle such questions because they can only handle data that is present in the training set. In recent years, structured knowledge representations have drawn much interest\cite{wu2017visual}. Large-scale Knowledge Bases (KBs) such as DBpedia \cite{auer2007dbpedia}, Freebase \cite{bollacker2008freebase}, YAGO \cite{hoffart2013yago2}, OpenIE\cite{etzioni2008open}, NELL \cite{carlson2010toward}, WebChild\cite{tandon2014webchild,tandon2014acquiring} and ConceptNet\cite{liu2004conceptnet} arose as a result of this. In a machine-readable format, these databases maintain common sense and factual information.. A fact is a piece of information that is frequently expressed as a triple (arg1, rel, arg2), having arg1 and arg2 representing two facts and rel signifying their relationship. 
	
	A method called Ahab \cite{wang2015explicit} for VQA, which has reasoning capabilities about an image on the basis of information extracted from an external knowledge base and constructs the KB-VQA dataset which contains 700 images and 2402 questions. Despite having a lower overall accuracy (69.6\% vs. 44.5\%) than joint embedding approaches, this model significantly outperforms them on the KB-VQA dataset [78]. In particular, Ahab significantly outperforms joint embedding methods on visual questions that guarantee a sufficient higher level of knowledge.
	
	FVQA (Fact-based VQA)\cite{wang2017fvqa}, a larger VQA dataset which requires and supports a deeper reasoning compared to KB-VQA \cite{wang2015explicit}. A novel method for answering questions with supporting-facts is also proposed. The process begins by detecting relevant information in the image and relating it to a knowledge base that has already been built. After that, the question is automatically identified and mapped to a query that searches the combined picture and knowledge base data. The query's response generates supporting facts, which are subsequently processed to provide the final answer. Although less manual processing is required compared with \cite{wang2015explicit}, \cite{wang2017fvqa} still has some weaknesses which hinder it from further development. Firstly, it only supports 13 relationships, which are obviously far from enough in open-ended VQA. Moreover, the answer is limited to the two concepts in the supporting fact, while there exists other formats of answers such as why, how, and how many, etc.
	Some approaches for improving FVQA\cite{wang2017fvqa} performance have been offered. Instead of using keyword matching techniques, \cite{narasimhan2018straight} provides a learning-based strategy that uses a learnt embedding space to get straight to the facts. In specifically, it learns a dynamic routing of statistics and question-image pairings to an embedding space. To answer a question, the fact that is most matched with the given question-image pair is chosen.By considering all entities together,\cite{narasimhan2018out} develops an entity graph and utilizes a graph convolutional network to reason about the definitive answer. The motivation is that the sequential method of forming a local decision by considering one fact at a time is inefficient.  \cite{narasimhan2018out} On the FVQA dataset, it performs admirably.
	\begin{figure}
		\centering
		\includegraphics[width=\linewidth]{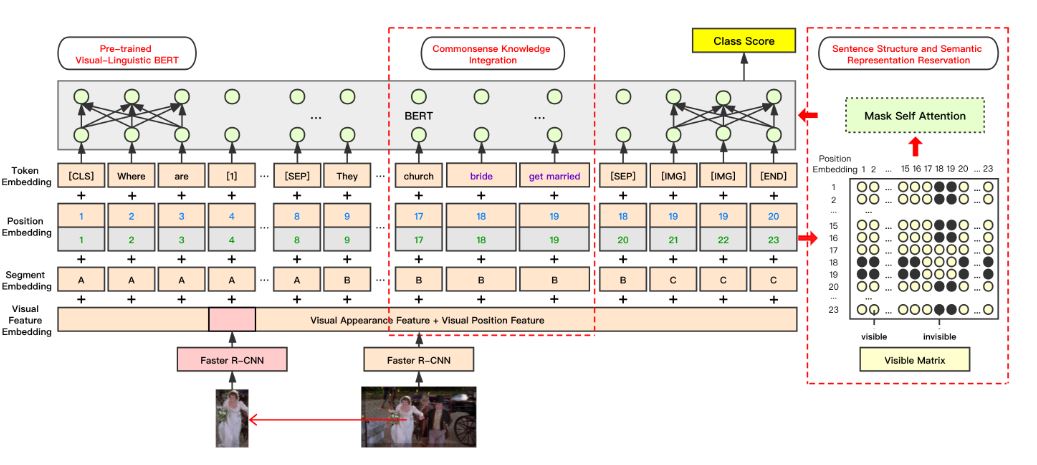}
		\caption{The architecture of KVL-BERT.\cite{song2021kvl}}
	\end{figure}
	\cite{shah2019kvqa} Contends that answering questions requiring world knowledge about named entities is also a major research challenge, in addition to the commonsense knowledge offered by KB-VQA and FVQA. \cite{shah2019kvqa}Provides KVQA, the first dataset for the job of (world) knowledge-aware VQA, to address this issue. KVQA has 183K question-answer pairings with over 18K named entities and 24K photos in total. The only dataset that identifies named entities and the need for knowledge about them is the KVQA dataset. The approaches discussed above take great care to create a knowledge base that includes answer-related triplets and use explicit reasoning to deduce answers. These approaches, on the other hand, are unable to scale to open-vocabulary datasets with a large and diversified query space. To this purpose, some approaches \cite{wu2016ask, wang2017vqa} use an external knowledge base in the feature extraction phase rather than the answer inference phase, while keeping the neural network as the backbone. This will enable more questions to be answered effectively, even if the image does not contain the entire response. \cite{wu2016ask} Provides an approach for answering visual questions that blends an internal representation of an image's content with information collected from a generic knowledge base to answer a wide range of image-based questions. In this method, image, caption, and knowledge representations are combined and input into an LSTM, which analyzes the question and generates the response. \cite{wang2017vqa} For better feature combining, a unique sequential co-attention model is used. Furthermore, the suggested method creates human-readable reasons for its decisions and may be trained end-to-end without the need for ground truth reasons.
	\begin{figure}
		\centering
		\includegraphics[width=\linewidth]{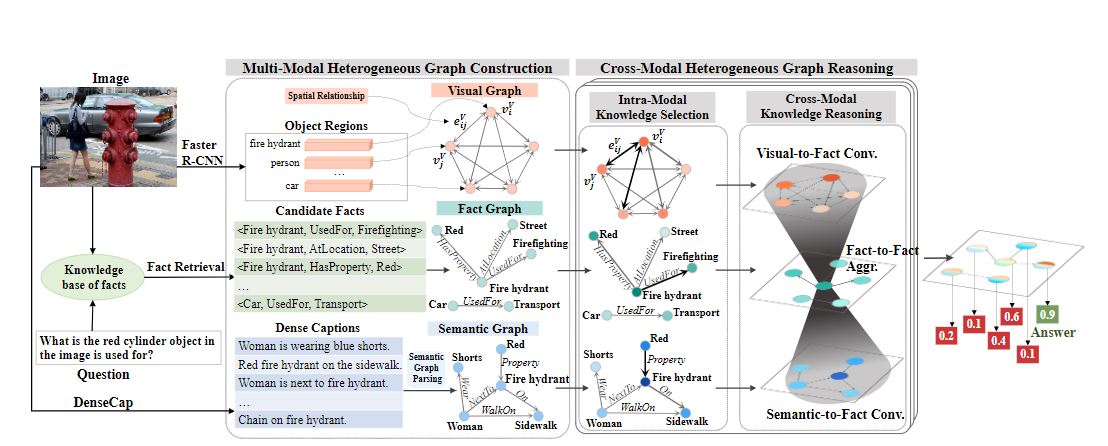}
		\caption{A high-level overview of Mucko. The model is made up of two modules:The idea of Multi-modal Heterogeneous Graph Construction is to build a representation of an image using various graph layers. The selection of intra-modal and cross-modal evidence is feasible using Cross-modal Hetegeneous Graph Reasoning.\cite{zhu2020mucko}}
	\end{figure}
	Existing FVQA [4] have the drawback of combining all datasets without fine-grained selection, which generates unexpected noises while justifying the final answer. Capturing question-oriented and knowledge evidence remains a significant challenge in resolving the issue. 
	
	Mucko \cite{zhu2020mucko}, a multi-modal heterogeneous graph with multiple types of layers correlating to visual, linguistic, and fact based features, and a modality-aware heterogeneous graph modeling technique on top of the multi-layer graph representations to obtain evidence from different layers that is most relevant to the given question, was proposed. 
	
	KVL-BERT \cite{song2021kvl} proposed a cross-modal BERT that incorporates commonsense knowledge. The multi-layer Transformer employs additional commonsense information that has been collected from ConceptNet\cite{liu2004conceptnet}. To retain the spatial features and lexical description of the initial phrase, the authors utilize relative parameter of the embedding and self attention to decrease the influence among the inserted commonsense knowledge and other irrelevant components in the input sequence.
	
	KRISP \cite{marino2021krisp} integrates symbolic representations from a knowledge graph with the robust implicit reasoning of transformer models for answer prediction, whilst not losing their explicit linguistic features to an implicit embedding. To address the tremendous variety of information required to answer knowledge-based questions, the model merges several sources of knowledge.
	\begin{figure}
		\centering
		\includegraphics[width=\linewidth]{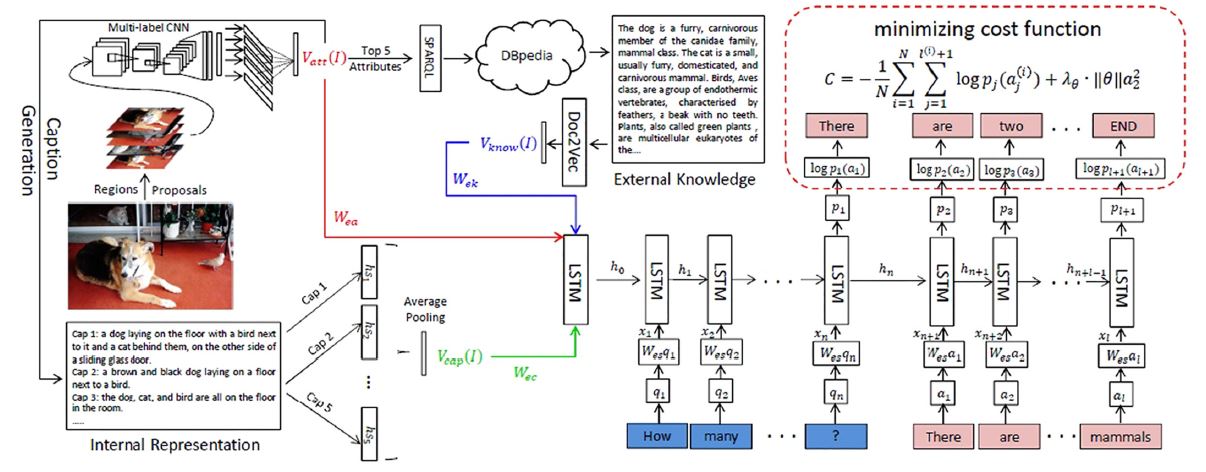}
		\caption{VQA model that use an external knowledge from DBpedia \cite{wu2016ask}}
	\end{figure}
	\subsection{Method based on Neural Network}
	\subsubsection{Traditional Neural Network Models}
	VQA is most commonly performed by sending both picture and text input to a single embedding. 
	
	These so-called joint embedding techniques, which were inspired by advances in deep learning for both NLP and computer vision, allowed the learning of representations in a global feature space. The first application of this technique was in the field of image captioning\cite{vinyals2015show}. It has been shown to be helpful for drawing conclusions from input that includes images and questions, making it a good fit for the VQA task.
	In its most basic form, the joint embedding approach can be separated into three steps. First, both the image and question features are represented in a useful way. These representations must then be merged in such a way that both visual and linguistic elements are preserved. This combined representation is then used to determine the response output in the last phase. One of the first attempts to tackle the VQA task was proposed by Malinowski and Fritz \cite{suhr2017corpus}. CNN Image features, along with the question, are fed into an LSTM, which then generates the answer.
	
	VIS+LSTM\cite{ren2015exploring}became the first study to view the VQA task as a classification problem over the answers. \cite{antol2015vqa}standardized the task by introducing the VQA v1.0 dataset \cite{agrawal2017c}, a performance assessment measure, and a yearly competition to stimulate research on the VQA task. The synchronizations of VGGNet's final hidden layer \cite{simonyan2014very} are utilized as image attributes in their baseline, deeper LSTM Q + norm I, which incorporates a 2-layer LSTM encoder. A simpler baseline was provided by \cite{radford2019language}. The question representation is obtained as a Bag of Words \cite{harris1954distributional} embedding individual question words in the CNN + BoW \cite{jabri2016revisiting} baseline model for VQA. GoogLeNet \cite{szegedy2015going} was used to obtain the visual attributes. In LSTM + CNN, The picture features are encoded using a CNN such as VGGNet \cite{gheller2018deep}, ResNet \cite{he2016deep}, and others in this baseline model \cite{malinowski2014multi}. For question representation, the final hidden state of LSTM \cite{suhr2017corpus} is utilized, which takes the question word embedding as input. To acquire the distribution over the collection of replies, the picture and question representations are concatenated and then put through an MLP. \cite{ren2015image} proposed a method for dealing with answers of varying word lengths in which the concatenation of picture and question feature vectors is fed into an LSTM encoder.
	
	Multi-Graph Reasoning and Fusion (MGRF) layer\cite{liu2014high} is a more recent study that leverages pre-trained contextual relational embeddings to reason sophisticated spatial and lexical relations between visual objects and adaptively fuse them. The MGRF layers may be built in depth to construct Deep Multimodal Reasoning and Fusion Networks to efficiently reason and fuse multimodal interactions (DMRFNet)\cite{zhang2021dmrfnet}. A tool for creating justifications is also added to verify the proposed answer. This rationale clarifies the decision-making motive of the model and increases its interpretability.
	
	N-KBSN\cite{ma2021joint}  is a joint embedding model based on dynamic word vectors, which differs from VQA techniques based on static word vectors. The authors of this paper divide the model into three major components: the question text and picture feature extraction module, the self-attention and directed attention module, and the feature fusion and classifier module. The BUTD technique extracts the visual features, and each image is characterized by a dynamic number (from 10 to 100) of 2048-D characteristics.
	
	\subsubsection{Attention Mechanism Models}
	A neural network's attention mechanism allows it to focus on select sections of the input that are more critical for addressing these task \cite{zhang2021dmrfnet}. Several attention models have been developed in the past in VQA and visual reasoning, allowing to learn attention distributions over visual, textual, or both modalities \cite{lu2016hierarchical,ben2017mutan,xu2016ask}. 
	
	The attention mechanism Stacked Attention Networks (SAN) was first postulated by\cite{yang2016stacked}, it has pushed the boundaries of neural network research to new heights. In the VQA community, the same thing happens. Several attention-based approaches have recently been proposed, all of which have greatly improved the performance of VQA models. Answering visual issues intuitively entails focusing on specific spatial regions while disregarding irrelevant objects. All of these models are based on the assumption that some visual regions in an image and specific phrases in a question seem to be more relevant for answering a given topic than others. \cite{yang2016stacked} First, by combining the spatial image features $v$ and query features, it produces attention over the picture pixels $v$
	
	\begin{figure}
		\centering
		\includegraphics[width=\linewidth]{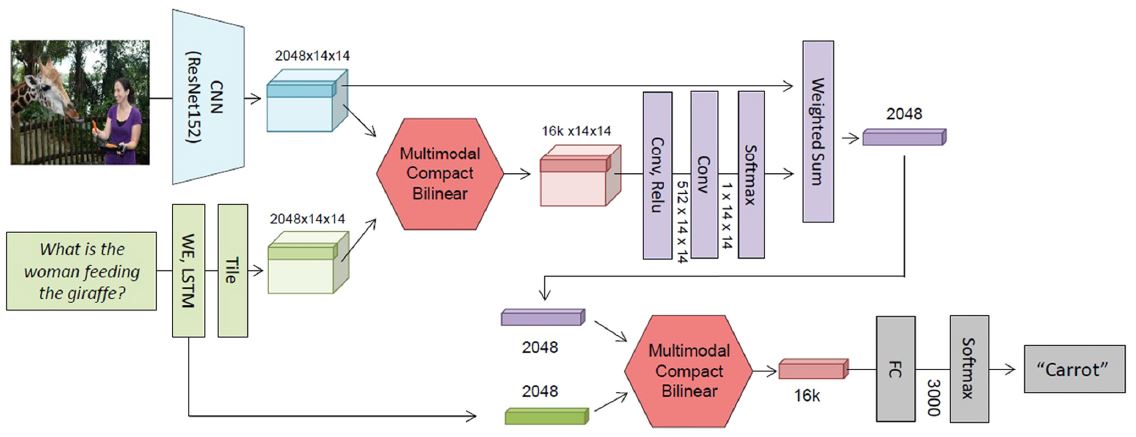}
		\caption{Architecture of Multi-modal Compact Bilinear (MCB) model with Attention mechanism }
	\end{figure}
	
	\begin{equation}
		\begin{aligned}
			h _{A} &=\tanh \left( W _{I, A} v_{I} \oplus\left( W _{Q, A} v_{Q}+ b _{A}\right)\right) \\
			p ^{I} &=\operatorname{softmax}\left( W _{P} h _{A}+ b _{P}\right)
		\end{aligned}
	\end{equation}
	The query vector $u$ is formed by adding a cumulative sum of visuals characteristics making use of the attention map (denoted by $\widetilde{ v }_{I}$ ) feature matrix in answer to the question vector $v_{Q}$.
	
	\begin{equation}
		\begin{aligned}
			\widetilde{ v }_{I} &=\sum_{i} p_{i}^{I} v _{i} \\
			u &=\widetilde{ v }_{I}+ v _{Q}
		\end{aligned}
	\end{equation}
	
	The above-mentioned attention process is repeated for $K$ attention layers. After the $k^{t h}$ attention step, the multi-modal space, the modified question vector $u^{k}$ signifies the fine-grained question.
	
	\begin{equation}
		\begin{aligned}
			h_{A}^{k} &=\tanh \left(W_{I, A}^{k} v_{I} \oplus\left(W_{Q, A}^{k} u^{k-1}+b_{A}^{k}\right)\right) \\
			p^{I, k} &=\operatorname{softmax}\left(W_{P}^{k} h_{A}^{k}+b_{P}^{k}\right) \\
			\tilde{v}_{I}^{k} &=\sum_{i} p_{i}^{I, k} v_{i} \\
			u ^{k} &=\widetilde{ v }_{I}^{k}+u^{k-1}
		\end{aligned}
	\end{equation}
	The classifier is fed the final question vector  $u^{k}$ to forecast the dispersion across the set of answers.
	
	The MAC network \cite{hudson2018compositional} was developed to help people reason more explicitly and expressively \cite{gurari2018vizwiz}. It is based on the usage of recurrent structure, internal memory, and attention mechanisms. The model, like RNN, is made up of a series of MAC cells linked together, each of which has two hidden states: control and memory. In contrast to standard neural networks, the MAC network features a distinct control and memory component. The control component of each MAC cell is in charge of recognizing the reasoning operation, while the memory component is in charge of storing the intermediate result. As a result, rather than approximating direct transformation between the input and output, the MAC network's structural prior enables the model to do iterative reasoning by breaking the problem into a succession of operations. According to the paper's findings, MAC networks can learn much faster and with significantly greater data efficiency than other approaches (using only 10\% of the data to achieve 85\% accuracy). Inductive bias can improve both learning accuracy and efficiency, according to this study. Furthermore, because the attention maps of the model might reveal how the model came up with an answer, MAC networks provide for better transparency and interpretability. Traditional deep learning models, on the other hand, are regarded a black-box system. However, the MAC networks' greatest flaw is that their performance on datasets containing real-world photos, such as GQA \cite{hudson2019gqa} isn't really challenging.
	
	The BUTD (Bottom-Up and Top-Down Attention for VQA) model \cite{anderson2018bottom} is based on the human visual system. They employed terminology from the neuroscience discipline in the paper, with top-down attention referring to focused and volitional attention driven by the current task, and bottom-up attention referring to automatic attention triggered by salient stimuli. The majority of VQA models (described in the preceding section) use top-down visual attention. The utilization of bottom-up attention is the main distinction between BUTD and the preceding approach (SAN). BUTD leverages image regions supplied by an object detector model (Faster R-CNN \cite{ren2016faster} ) to express image features as bottom-up attention. This model outperformed other cutting-edge methods by a significant margin across all question types in the 2017 VQA Challenge, achieved an overall score of 70.34\% with 30 ensembled models on the VQA v2.0 test-std set. Additionally, since its creation, this approach has been the most frequently used fundamental basis for VQA research. However, the effectiveness of the VQA models is influenced by the Faster R-CNN object detector's capacity, and a larger capacity can aid retrieve more expressive significant findings.
	
	The Neuro-Symbolic Concept Learner (NS-CL) \cite{mao2019neuro} is a framework based on the idea of neuro-symbolic AI, although all prior models, which are based on completely differentiable neural networks. The goal of the neural-symbolic method is to combine symbolic AI's reasoning capacity with neural networks' learning capabilities \cite{garcez2019neural}. Comparable to human concept learning, NS-CL can gain knowledge on visual concepts, words, and semantic parser of text without explicit supervision. In the NS-CL framework, DNNs are used for image recognition and language understanding, while symbolic program execution is used to handle the reasoning aspect.
	
	Recently, \cite{hong2020selective} proposed model called Selective Residual Learning (SelRes) that makes use of the self-attention mechanism. To capture the most important relationships, the model employs residual learning. Selective masking was also proposed as a strategic approach for attention mapping depending on the relevance of the preceding stack's vector.
	\begin{figure}
		\centering
		\includegraphics[width=\linewidth]{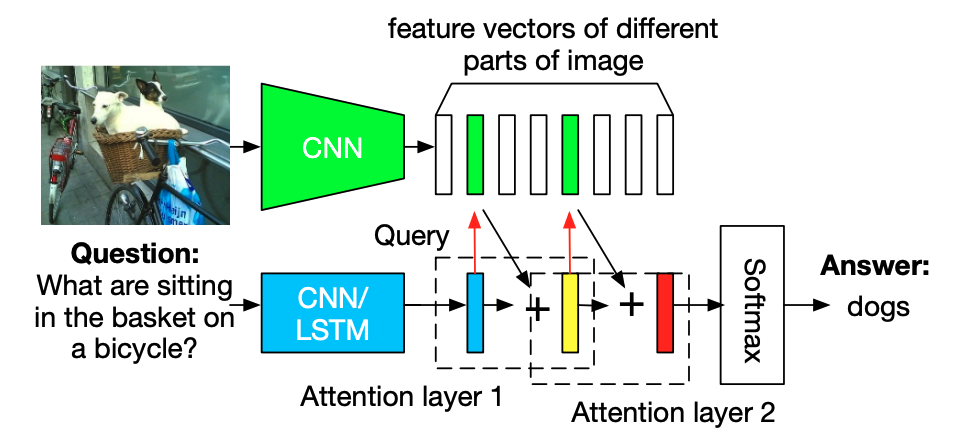}
		\caption{Overall architecture\cite{yang2016stacked} of Stacked Attention Networks}
	\end{figure}
	\subsubsection{Bilinear pooling Models}
	Bilinear pooling techniques have been researched as alternatives to element-wise products and simple concatenation for fusion. The attention mechanism and the bilinear pooling technique can both benefit from one another.
	
	The well know method for bilinear pooling method MCB (Multimodal Compact Bilinear Pooling) \cite{fukui2016multimodal} is a multi modal compact bilinear pooling technique.The picture and question parameters are derived via LSTM and CNN, respectively, in this model, as in prior models. To integrate the two modalities, it uses Multimodal Compact Bilinear pooling (MCB) \cite{gao2016compact}. The integration takes place at each position in the spatial grid of features generated by the CNN. The score for soft attention of each place is then predicted using two convolutional layers followed by softmax. Using the attention map, the weighted sum of visual features is merged with the question understanding using MCB once again. To acquire the distribution over the set of replies, the features are fed via an MLP and then softmax. A Fast Fourier transformation is used to transform the spectral domain (FFT).
	\begin{equation}
		z= FFT ^{-1}\left( FFT \left(v^{\prime}\right) \odot FFT \left(q^{\prime}\right)\right)
	\end{equation}
	In this method, the frequency domain product is translated back to the original domain via an inverse Fourier transformation, where convolution in the time-frequency domain is the same as element-wise sum in the time-frequency domain. However, the model is still somewhat costly to train since, in order to achieve great VQA accuracy, the results in a change of joint embedding vector $z$ must be close to the edge (exactly 16, 000d).
	
	The achievement of bilinear encoding techniques like MCB and MLB in VQA tasks is noteworthy.On the VQA v1.0 real test-std set for open-ended questions, MCB in particular achieved cutting-edge success with an overall score of 66.5\%. Despite having considerably fewer computational parameters, MLB managed to achieve a challenging score of 66.89\%. The high cost of computation is the biggest drawback of bilinear encoding techniques.
	
	Furthermore, Bilinear Attention Networks (BAN)
	\cite{zhu2016visual7w} concurrently learn textual and visual attention, resulting in a mapping from the image's recognized objects to the question's words. Bilinear Graph Networks \cite{guo2021bilinear} takes it a step further by understanding BAN as a graph, allowing it to construct associations between objects and words via the image graph and between words and words via the question graph.
	
	A low-rank bilinear pooling\cite{kim2016hadamard} minimize the output feature vector's dimensions to replace compact bilinear pooling, which requires expensive computations and has a fan-out structure. When compared to compact bilinear pooling, low-rank bilinear pooling has a more flexible structure that uses linear mapping and the Hadamard product, as well as a superior parsimonious property. In general, the VQA model created thus far is more accurate than the previous MCB technique. Despite the fact that MLB and MCB have comparable performance, it requires more time for them to merge. 
	
	Multi-modal Factorized Bilinear Pooling (MFB)\cite{yu2018beyond} is utilised to further fuse the attended multi-modal features, by adding a pooling procedure to the jointly embedded feature vector.The MFB approach is developed to more efficiently combine the textual features from the question and the visual features from the image. By allowing for more effective utilization of the complex correlations between multi-modal features, this approach can significantly outperform the existing bilinear pooling approaches.
	Multiple MFB blocks are cascaded to establish a generalized Multi-modal Factorized High-order pooling (MFH)\cite{yu2018beyond} method. In contrast to MFB, MFH captures more complicated correlations of multi-modal features to achieve more racially biased embeddings, which further leads to a notable improvement in the VQA efficiency.
	
	Multimodal Tucker Fusion (Mutan) \cite{ben2017mutan} was the first VQA model to use tensor decomposition algorithms to reduce the dimensionality of input visual and textual feature vectors, as well as the output joint feature embedding. A block-super diagonal tensor \cite{ben2019block}  based decomposition (Block) technique for VQA to address this bottleneck challenge. 
	Recently, \cite{vu2020question} propose a  model for medical imaging called Question Centric Multimodal Low-rank Bilinear (QC-MLB). The model join question and visual elements that are begin organized. They also shows that proposed VQA model may be used in conjunction with CAM-like approaches to acknowledge which portions of the image the model uses to anticipate the answer.
	\subsubsection{Transformer models}
	MCAN (Deep Modular Co-Attention Networks) is a VQA model based on the Transformer \cite{yu2019deep} architecture. It makes use of the Transformer model's self-attention and guided-attention units. MCA, which is the model's core element, is made up of these two fundamental elements.Intense intra-modal and inter-modal relationships are made possible by the MCA layer. The self-attention unit enables the model to learn relationships between words in questions and between regions in images. However, using the guide-attention unit, the image-question relation is simpler to understand. The MCA layer can also be layered deeper for more precise visual reasoning. To extract the question features, GloVe word embedding \cite{pennington2014glove} is used to convert the words in the question into vectors. Then, to retrieve the contextual representation of each word, word vectors are given to LSTM.
	
	On the VQA-v1 test-dev set, the neural module network (NMN) model performs better than other cutting-edge models, achieving a total score of 58.0\%. Answers that refer to an object, an attribute, or a number perform particularly well for NMN. However, using an improved parser or using joint learning can dramatically reduce parser errors, improving how well VQA tasks perform.
	\begin{figure}
		\centering
		\includegraphics[width=\linewidth]{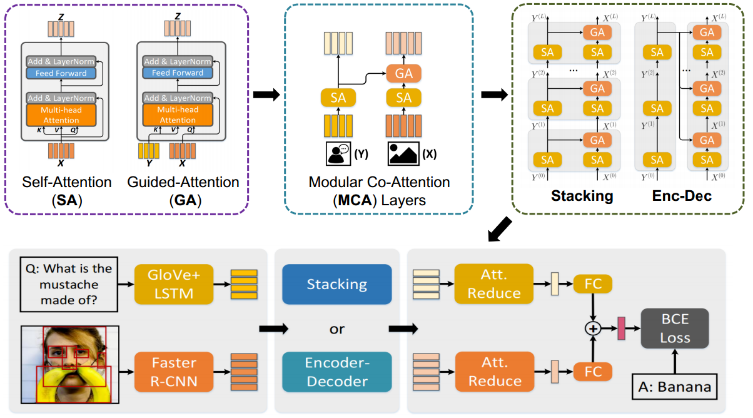}
		\caption{ Overall architecture of Deep Modular Co-Attention Networks \cite{yu2019deep}}
	\end{figure}
	
	Pixel-BERT \cite{huang2020pixel} present a Visual Encoder based on CNN that is combined with several modal Transformers to build Pixel-BERT from start to finish and generate more accurate and thorough embedding between visual and linguistic contents at the pixel and text level. For the robustness of visual embedding learning, we take pixels from an image as input and apply a random pixel sampling technique. On the Visual Genome\cite{krishna2017visual} dataset and the MSCOCO dataset\cite{chen2019hybrid}, they built a pre-training model based on Pixel-BERT to learn a universal visual and linguistic embedding. Two pre-training tasks are the masked language model and image-text matching.
	
	In the Self-Adaptive Neural Module Transformer SANMT\cite{zhong2020self} model, question features are encoded using a novel encoder-decoder architecture that feeds forward, and a transformer module is used to produce a dynamic question feature embedding that increases in size as the reasoning processes advance. the model is also used the intermediate findings to dynamically alter the layout choices and choose the question features.
	
	\subsection{Method based on Explicit Reasoning}
	The disadvantage of using CNNs or RNNs for visual or cognitive feature extraction is that these extraction models are monolithic by nature, which is a limitation of these techniques. Joint embedding techniques employ a "one size fits all" method of VQA, where the model applies the same computations to each input regardless of the type or degree of difficulty of the question. Utilizing a modular approach, whose size and shape are determined by the nature of the inputs, would make sense. In this section, the modular compositional models are briefly discussed.
	
	\subsubsection{Modular Neural Models}
	Neural Module Networks (NMN)\cite{andreas2016neural} was the first to be developed and \cite{andreas2016learning} makes an improvement on the model. It's a strategy for addressing complicated questions that involves spatial reasoning and the capacity to identify shape and object properties. Each module is a neural network with a predefined layer structure and parameters that are learned during training. Stanford Parser \cite{ren2015exploring}is used to obtain the universal dependency parse representation. The key feature of this methodology is that it effectively employs the syntax tree to adapt well to complicated compositional reasoning challenges. The layout predictor has been significantly custom for this unique dataset and must be re-specified for usage on other datasets, which is a drawback. 
	
	N2NMN \cite{hu2017learning} is an acronym for End-to-End Module Networks for VQA.
	In a myriad of different ways, this model can improve (NMNs) method\cite{bahdanau2014neural}. The layout predictor is a decision network with a sequential RNN that generates a series of tokens ($\left\{m^{(t)}\right\}$) equivalent to the Reversible Polish module syntax tree notation\cite{burks1954analysis}. An LSTM encoder is used to encode the input question (which consists of $T$ words) to create a collection of encoder states $\left[h_{1}, h _{2}, \cdots, h _{T}\right]$. For each utterance in the training data, the decoder predicts attention weights at timestamp t.
	\begin{equation}
		\begin{aligned}
			u_{t i} &= v ^{T} \tanh \left( W _{1} h _{i}+ W _{ 2 } h _{t}^{\prime}\right) \\
			\alpha_{t i} &=\frac{\exp \left(u_{t i}\right)}{\sum_{j=1}^{T} \exp \left(u_{t j}\right)} \\
			c _{t} &=\sum_{i=1}^{T} \alpha_{t i} h _{i}
		\end{aligned}
	\end{equation}
	The hidden state of the decoder LSTM at time step $t$ is denoted by $h _{t}^{\prime}$, the context vector is denoted by $c _{t}$, and the question representation is denoted by $q \cdot v, W _{1}$, and $W _{2}$ are parameter tensors that can be learned. 
	
	Dynamic Memory Networks (DMN)\cite{kumar2016ask}, is another modular technique for question answering which combines a memory component with attention mechanisms  This approach was initially used on the VQA task by Xiong et al. \cite{xiong2016dynamic}. Their model includes a phrase input method that are join in natural language input and send it to a bi-directional GRU layer that extracts a collection of "facts" from the input. The GRU's bi-directional nature allows both past and future time steps to influence its internal state at a given time step, allowing information to flow between distinct inputs. The GRU's bi-directional features allow its internal state at a given time step to be changed by both current and future time steps, allowing information to flow between distinct inputs.
	On the VQA v1.0 test-std set, the dynamic neural module network (D-NMN) technique obtains an overall score of 58.0\% with strong understandability. Submodules must be thoughtfully designed, though, by the D-NMN. It is not possible to adapt these predefined modules to other datasets. The implementation of neural module networks is thus still an issue.
	
	In contrast to the original NMN \cite{andreas2016neural}, which relied on hand-crafted off-the-shelf syntactic parsers, \cite{johnson2017inferring} uses reinforcement learning to teach a program generator to adapt to progressively difficult problems. The program generator and the execution engine are built from neural networks trained by combining backpropagation and REINFORCE. The fundamental disadvantage of \cite{johnson2017inferring} is that each instantiation of a functional module corresponds to a separate neural network, leading to a large training space as the size of the evaluation dataset increases. This is why \cite{johnson2017inferring} struggles to grow to datasets with open-vocabulary terms. The authors Solely reports findings from the CLEVR \cite{johnson2017clevr} closed-vocabulary dataset.
	The NMN techniques outperformed other cutting-edge techniques on the VQA v1.0 test-dev set, scoring a total of 58.0\%. The performance of NMN is extremely good when questions are answered in terms of an object, attribute, or number. To improve the performance of VQA tasks, a better parser or the use of joint embedding can help reduce parser errors.
	
	Stack-NMN \cite{hu2018explainable} was developed to overcomes one of the primary drawbacks of prior techniques in this regard \cite{andreas2014much,hu2017learning}, which required hand-engineering the layout or learning it through a policy gradient. It’s a soft and continuous module structure that performs well even without layout annotations. The most essential advantage of Stack-NMN is that it uses a differentiable stack structure to enable soft layout selection, making the execution mechanism fully differentiable so that the model can be trained with back-propagation like other neural networks. 
	The VQA language bias concern the future of how VQA models are prone to learning solely from question-and-answer pairs without comprehending visual content. 
	The Transparency by Design network \cite{mascharka2018transparency} is based on the End-to-End Module Networks \cite{hu2017learning}, however there is one significant difference:instead of learning whether or not to use attention maps as inputs, the suggested modules directly use them. Because the modules perform specified functions, this makes them more interpretable. This model achieved state of the art On the CLEVR CoGenT dataset, \cite{johnson2017clevr}.
	
	Memory-augmented networks MANs \cite{ma2018visual}, which draw inspiration from co-attention mechanisms and memory-augmented neural networks.
	The MAN methods use co-attention to jointly embed image and question features in an abided memory-augmented network in order to identify sparse exemplars in training data. The structure of model is clearly different from that of the DMN, includes both an internal memory located inside LSTM and an external memory that is under the control of LSTM. The input images are processed to extract image features using the pre-trained VGGNet-16 and ResNet-101 of the MAN. The embedded word tokens are used as the input for the bidirectional LSTMs to build fixed-length sequential word vectors. The features that are most significant to each modality in relation to the other modality are then highlighted using a sequential co-attention mechanism based on the image and question features.
	
	On the VQA v1.0 and VQA v2.0 test sets, the memory-augmented networks (MAN) method outperforms the state-of-the-art MCB method in a competitive manner. On the VQA v1.0 dataset, the MAN performs marginally better on multiple-choice questions and marginally worse on open-ended questions. The MAN outperforms the DMN+ by a significant margin of 3.5\% because it only utilizes the internal memory of RNNs rather than an enhanced external memory. The MAN performance only trails the MCB by about 0.2\% on the VQA v2.0 test set.
	
	Feature-wise Linear Modulation (FiLM) \cite{perez2018film} is a general-purpose conditioning mechanism described in a recent approach. As affine functions of the GRU \cite{sridharan2019amrita} representation of the question, FiLM learns the Batch Normalization parameters for each channel.
	\begin{equation}
		\begin{aligned}
			\gamma_{i, c} &=f_{c}\left( x _{i}\right) \\
			\beta_{i, c} &=h_{c}\left( x _{i}\right) \\
			F i L M\left( F _{i, c} \mid \gamma_{i, c}, \beta_{i, c}\right) &=\gamma_{i, c} F _{i, c}+\beta_{i, c}
		\end{aligned}
	\end{equation}
	The question representation for the $i^{t h}$ input example is denoted by $x _{i}$, while the BN metrics for the $i^{t h}$ input and $c^{t h}$ channel are denoted by $(\gamma, \beta)$. A CNN (either pretrained or trained from the beginning) is followed by four residual blocks in the model architecture. FiLM controls the parameters of each residual block, which comprises of convolutional and a batch-normalization layer. The classifier (which consists of a 2-layer MLP followed by softmax) outputs a distribution over the answers based on the features produced by the FiLM network. By explicitly translating the query into an explainable module assembly, NMNs effectively preclude the language-to-reasoning shortcut \cite{shi2019explainable}. However, as this model points out, the vision-to-reasoning shortcut remains a roadblock on the way to true reasoning. Explicit reasoning on visual interpretation is also required to address this. 
	
	In another research paper \cite{yi2018neural} goes a step farther in the direction of explicit reasoning by executing programs in pure symbolic space. It is assumed that running programs in a symbolic space is more resistant to long program traces. However, it is not applicable to the real-world issue because it relies on a clean image and a fixed symbol vocabulary. Another paper\cite{hudson2019learning} Provides a similar concept to NMN, in which the query is translated into a series of soft instructions and then sequential reasoning is performed across the scene graph. 
	There are also approaches that use explicit reasoning to comprehend images without relying on the NMN framework. \cite{teney2017graph} Recommends that scene items be used to create graphs, which are subsequently explored using a deep neural network. The scene graph is created solely for the purpose of fine-tuning features. 
	
	Entity Attribute Graph (EAG)\cite{xiong2019visual} proposes a new format for representing images that is similar to the traditional scene graph. After the scene graph and query graph have been produced, further graph matching is undertaken to retrieve the answer. \cite{aditya2018explicit} Transforms images and questions into triples and infers the most likely answer using Probability Soft Logic (PSL). However, approaches based on statistics, such as \cite{aditya2018explicit}, are dataset-dependent and thus difficult to scale. \cite{xiong2019visual} proposed a graph matching-based architecture in which a graph matching module finds the best match of the query graph in the scene graph once the scene graph and query graph are formed. The entire procedure is easily explained. The graph matching algorithm in \cite{xiong2019visual} has the problem of not performing well on open-vocabulary datasets.
	\begin{figure}
		\centering
		\includegraphics[width=\linewidth]{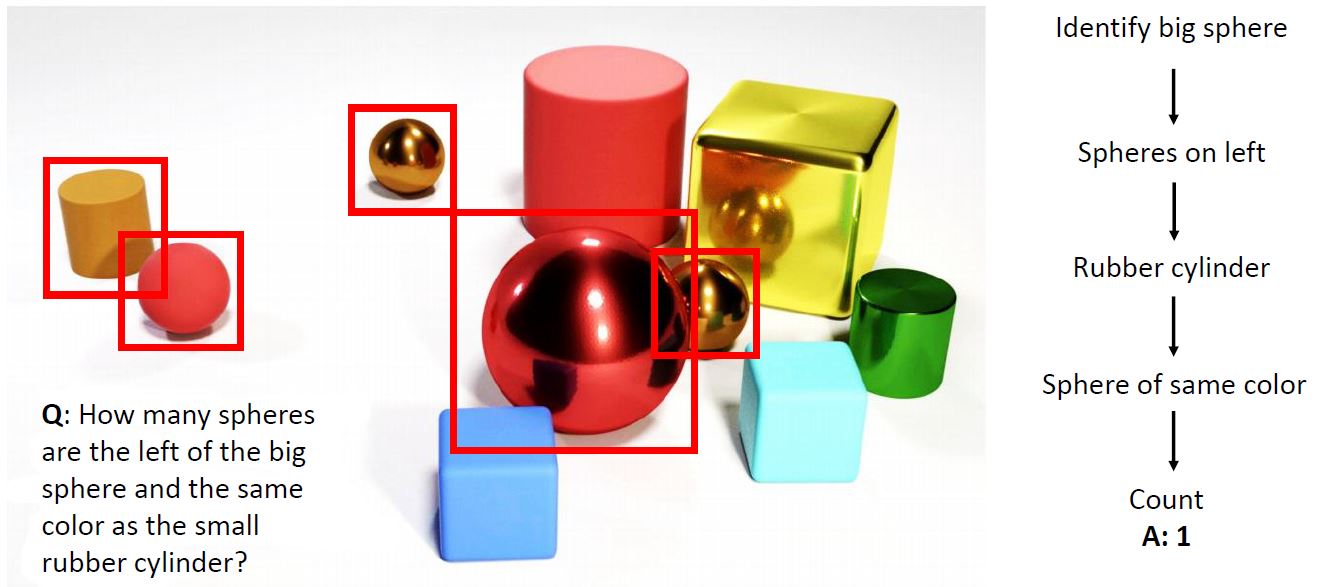}
		\caption{ Overall architecture of Deep Modular Co-Attention Networks \cite{yu2019deep}}
	\end{figure}
	\subsection{Other Approaches}
	\subsubsection{Graph and Neural-Symbolic}
	LRTA\cite{liang2020lrta}[Look, Read, Think, Answer], is a clear neural-symbolic reasoning model for VQA that improves performance step by step, precisely like humans, and offers human-readable explanation at each step. LRTA trains to parse a question into numerous reasoning steps after converting an image into a scene graph. The reasoning steps are then executed one after another exploring the scene graph with a recurrent neural-symbolic computation unit. At the end, it provides a more complete answer to the question with human language explanations. 
	
	To answer Visual reasoning tasks,\cite{guo2021bilinear}proposed bilinear graph networks. From a new graph perception, the model tries to analyze the bilinear attention map in both words in the question and objects in the image, then highlight the significance of using the intra-modality approach to language in the question and considering the cross-modality relationship between the question and image for complex reasoning. 

	A recent paper,\cite{hildebrandt2020scene}proposes a unique strategy for solving the challenge by undertaking context-driven, sequential reasoning based on the elements in the scene and their linguistic and geometrical relationships. As a preliminary step, the authors of this work generate a scene graph that characterizes the elements in the image, including their properties and mutual interactions. Subsequently, a reinforcement agent learns to walk autonomously across the extracted scene graph to construct pathways, which are then utilized to deduce answers.
	
	The interpretability of VQA modeling techniques, which are frequently used in deep learning models, is extremely difficult. An object-difference-based graph learner that focuses on teaching question-adaptive semantic relations is proposed by \cite{zhu2020object} as a solution to these issues by calculating inter-object differences while being guided by questions. The learned relationships enable the output image to be transformed into an object graph with structural dependencies between the objects. On the VQA-v2 test set, the graph learner method receives a total score of 66.18\%, which is comparable to the current approaches. Qualitative results highlight the proposed model's simplicity and clarity. Since the more intricate connections between graph items cannot be managed by the straightforward graph structure used in the proposed model, a more intricate architecture can be used to do so.
	
	The "Linguistically driven Graph Capsule Network" \cite{cao2020linguistically} is a hierarchical compositional reasoning paradigm in which the linguistic parse tree guides the composing process. The authors combine the lexical embedding of one word in the initial question with visual proof by integrating each capsule in the bottom layer. If related words are discovered in the parse tree, they are then directed to the same capsule.
	
	The Relation-aware Graph Attention Network (ReGAT) \cite{li2019relation}simulates various types of inter-object interactions by encrypting each image as a graph and using a graph attention mechanism. Multimodal Graph Memory Neural Networks. The relation-aware graph attention network (ReGAT) model outperforms cutting-edge methods on the VQA v2.0 test set, scoring 70.58\% overall. Due to its compatibility with conventional VQA models, the ReGAT model can be quickly incorporated into existing VQA models. The introduction of ReGAT has significantly improved the performance of contemporary VQA models on the VQA v2.0 val set.

	MN-GMN \cite{khademi2020multimodal}, a new neural network architecture for VQA. MN-GMN uses a graph structure and the recently proposed Graph Network (GN) in order to make decisions about the objects and how they interact with one another in a scene. There are four modules that make up the MN-GMN. The input module creates a set of visual feature vectors and a set of captions that are region-grounded and encoded for the image (RGCs).

	\subsubsection{Diagnosis Method}
	Agrawal et al. (2018) propose a GVQA\cite{agrawal2018don} with architectural constraints and inductive biases that are specifically intended to prevent the model from cheating during training by relying only on priors. The model, in particular, distinguishes between identifying a feasible answer space for a given question and identifying visual features in an image, allowing the model to generalize more robustly across a range of answer options.
	According to recent research \cite{wang2021latent}, latent variable models for VQA are recommended, in which extra information (like captions and answer categories) is included as latent variables that are observed during training but enhance question-answering performance at test time. An end-to-end architecture that generates a multi-word answer to a visual question was proposed by Dua et al. in 2021\cite{dua2021beyond}.

	\subsubsection{Memory-Based}
	Su et al., 2018 proposed a visual knowledge memory network (VKMN) \cite{su2018learning}, which seamlessly combines structured human knowledge and deep visual elements into memory networks in a learning framework from start to finish. The first is a technique for mixing visual content with information about knowledge. VKMN solves this issue by combining knowledge triples (subject, relation, and target) and deep visual characteristics. The second method handles numerous knowledge facts obtained from question and answer pairings. CM-VQA \cite{jiang2015compositional} is as an end-to-end training framework that explicitly fuses the features associated with words and those accessible at different local patches in an attention mechanism, then combines the fused information to construct dynamic messages, which the authors term as episodes. The episodes, together with the contextual visual and linguistic information, are then fed into a typical question-answering module.
	
	\section{Discussion and Future Direction }
	Vision and language integration has advanced significantly since the beginning of the research, particularly with the aid of deep learning methods. Although there is still a long way to go before SOTA models can match human talent, the gap is steadily closing. However, theory and computational advancements are still possible. Here, we offer a number of potential future directions that might aid in the advancement of the study as a whole.
	\subsection{Issues of datasets biases}
	The VQA and visual reasoning datasets that are currently available are biased. The question, rather than the image content, is used in the majority of SOTA VQA and Visual reasoning models. The way a question is phrased has a big influence on the answer. As a result, it will be far more difficult to evaluate VQA methods. Furthermore, because the bulk of these questions are strongly associated to the inclusion of objects or scene aspects, answering questions that entail the use of visual material is fairly simple. CNNs are qualified to respond to these questions. They also have a lot of linguistic prejudices. Questions that begin with the word "why" are more difficult to respond to and are less common. This will have a significant impact on performance appraisals.
	\subsection{Issues of Evaluations of opened-ended and multiple-choice questions}
	The majority of cutting-edge (SOTA) VQA systems can evaluate multiple-choice questions using traditional accuracy metrics. However, these systems are incapable of evaluating the open-ended question.When assessing M-C questions, rather of describing the question's solution, the task is simplified to just choosing the most plausible answer. In the case of M-C questions, the options must be provided to the system and ensure that perhaps the system is forced to analyze the content of an image instead of just assuming the answer from the available options.
	\subsection{Issues of Large Datasets Limitation}
	Most techniques developed for tasks that mix vision and language utilize huge datasets for training. With this tendency, designing new challenges in the absence of a dataset will become increasingly challenging. To prevent these problems in the future, work must be adaptive to datasets of varied sizes.
	\subsection{Issues of External knowledge}
	The VQA goal is a full machine intelligence task that includes image and language data to produce insightful question and answer. Moreover, the existing VQA aim is to discover the right answer from the predefined answer. One form of linguistic bias is the unequal distribution of information and question answers. As a result, rather than employing fixed question and answer types, a decent VQA method should seek for the right answers more extensively from the external data source. This severely limits the use of VQA. Utilizing full external data sets or merging knowledge maps to answer questions may be an attempt to eliminate language bias, allowing the model to produce a more equitably shared question answer dataset. One of the key directions to help reduce linguistic bias is how to identify more effective external knowledge aid solutions.
	\section{Conclusion}
	VQA is a difficult research topic that demands algorithms capable of complicated recognition and reasoning tasks. It leverages on recent breakthroughs in both computer vision and NLP approaches.	The field of artificial intelligence would advance significantly if an algorithm could solve all of the challenges raised by this task and answer any question about any image. An overview of relevant datasets frequently used to measure the effectiveness of VQA and visual reasoning algorithms is provided after an introduction to various computer vision and NLP tasks. These datasets contained natural and artificial images and all their annotations, such as supplementary information about the external setting required to answer specific visual questions. Then, we introduced a number of algorithms and divided them into several categories according to their main contributions. Instead of using models that perform well on certain datasets but poorly on others, future research would make use of more comprehensive models capable of addressing problems that call for knowledge of the outside world, complex reasoning, and trying to handle synthetic data.
	
	\bibliography{IEEEtran}
	\bibliographystyle{IEEEtran}
	
\end{document}